# Scheduling Bipartite Tournaments to Minimize Total Travel Distance

**Richard Hoshino**　　　　　　　　　　　　　　　　　　　　　　　richard.hoshino@gmail.com
**Ken-ichi Kawarabayashi**　　　　　　　　　　　　　　　　　　　　　k_keniti@nii.ac.jp
*National Institute of Informatics,*
*2-1-2 Hitotsubashi, Chiyoda-ku, Tokyo 101-8430, Japan*

## Abstract

In many professional sports leagues, teams from opposing leagues/conferences compete against one another, playing *inter-league* games. This is an example of a *bipartite* tournament. In this paper, we consider the problem of reducing the total travel distance of bipartite tournaments, by analyzing inter-league scheduling from the perspective of discrete optimization. This research has natural applications to sports scheduling, especially for leagues such as the National Basketball Association (NBA) where teams must travel long distances across North America to play all their games, thus consuming much time, money, and greenhouse gas emissions.

We introduce the *Bipartite Traveling Tournament Problem (BTTP)*, the inter-league variant of the well-studied Traveling Tournament Problem. We prove that the $2n$-team BTTP is NP-complete, but for small values of $n$, a distance-optimal inter-league schedule can be generated from an algorithm based on minimum-weight 4-cycle-covers. We apply our theoretical results to the 12-team Nippon Professional Baseball (NPB) league in Japan, producing a provably-optimal schedule requiring 42950 kilometres of total team travel, a 16% reduction compared to the actual distance traveled by these teams during the 2010 NPB season. We also develop a nearly-optimal inter-league tournament for the 30-team NBA league, just 3.8% higher than the trivial theoretical lower bound.

## 1. Introduction

Consider a tournament involving two teams $X$ and $Y$, each with $n$ players. In a *bipartite tournament*, players from team $X$ compete against players from team $Y$, with the goal of determining the superior team. Labeling the players $\{x_1, x_2, \ldots, x_n\}$ and $\{y_1, y_2, \ldots, y_n\}$, we represent each match by the ordered pair $(x_i, y_j)$, with indices $i, j \in \{1, 2, \ldots, n\}$.

The Davis Cup is an example of a bipartite tournament, where each country fields a tennis squad consisting of two singles players and a doubles team. Their are five matches played between the two countries, with the doubles teams squaring off on Day 2, sandwiched between the singles matches $(x_1, y_1), (x_2, y_2)$ on Day 1, and $(x_1, y_2), (x_2, y_1)$ on Day 3. Another example is the biennial Ryder Cup championship, where the United States and Europe field teams consisting of the top twelve male golfers. The competition culminates with twelve head-to-head matches on the last day, with the $i^{\text{th}}$ ranked golfer from the United States facing off against the $i^{\text{th}}$ ranked golfer from Europe.

In a single round-robin (SRR) bipartite tournament, each player from $X$ competes against every player from $Y$ once, with everyone playing one match in each time slot. This produces a tournament with $n^2$ matches spread out over $n$ time slots. In a double round-robin (DRR) bipartite tournament, each pair plays twice, thus producing a tournament with $2n^2$ matches spread out over $2n$ time slots. SRR bipartite tournaments are common in tennis and ping-pong, while DRR bipartite tournaments are used in chess, so that $x_i$ plays against each $y_j$ twice, with one game as white and one game as black. The aforementioned Ryder Cup is an example of a *partial* bipartite tournament, where each player from $X$ plays against some proper subset of players from $Y$.

While there has been much research conducted on the theory of bipartite tournaments (Kendall, Knust, Ribeiro, & Urrutia, 2010), all previous papers have dealt with feasibility and fairness, specifically in constructing balanced tournament designs and minimizing carry-over effects (Easton, Nemhauser, & Trick, 2004) to ensure competitive balance for all the players on each team.

By replacing the words "team" and "player" by "league" and "team", respectively, we can view $X$ and $Y$ as two $n$-team sports leagues, where a bipartite tournament between $X$ and $Y$ represents *inter-league play*. For example, Major League Baseball (MLB) holds four weeks of inter-league games each season, with every American League team playing 18 games against a half-dozen teams from the National League. MLB inter-league play is an example of a partial bipartite tournament, where some/many of the scheduled games are based on historical rivalry or geographic proximity.

In this light, we consider the problem of minimizing the *total travel distance* of bipartite tournaments. For chess and tennis, the issue of travel is irrelevant as all tournament matches take place in the same venue. However, in the case of inter-league play in professional baseball, teams must travel long distances to play their games all across North America, and so finding a schedule that reduces total travel distance is important, for both economic and environmental reasons.

To answer this question of creating a distance-optimal inter-league schedule, we introduce a variant of the *Traveling Tournament Problem* (TTP), in which every pair of teams plays twice, with one game at each team's home stadium. The output is an optimal schedule that minimizes the sum total of distances traveled by the teams as they move from city to city, subject to several natural constraints that ensure balance and fairness. Unlike the TTP which models a double round-robin *intra*-league tournament, our variant, the *Bipartite Traveling Tournament Problem (BTTP)*, seeks the best possible double round-robin *inter*-league tournament.

Since its introduction (Easton, Nemhauser, & Trick, 2001), the TTP has emerged as a popular area of study within the operations research community (Kendall et al., 2010) due to its incredible complexity, where challenging benchmark problems remain unsolved. Research on the TTP has led to the development of powerful techniques in integer programming, constraint programming, as well as advanced heuristics such as simulated annealing (Anagnostopoulos, Michel, Hentenryck, & Vergados, 2006) and hill-climbing (Lim, Rodrigues, & Zhang, 2006). More importantly, the TTP has direct applications to scheduling optimization, and can aid professional sports leagues as they make their regular season schedules more efficient, saving time and money, as well as reducing greenhouse gas emissions.

The purpose of this paper is to consider the problem of creating distance-optimal inter-league tournaments, thus connecting the techniques and methods of sports scheduling to the theory of bipartite tournaments, producing new directions for research in scheduling op-timization. Optimizing inter-league tournaments is a natural next step in the field of sports scheduling, especially since the introduction of inter-league play to professional sports. For example, in Major League Baseball, inter-league play began only in 1997, six decades after it was first proposed. In Japan, the Nippon Professional Baseball (NPB) league was formed in 1950, yet NPB inter-league play did not commence until 2005.

The authors were motivated to analyze the Japanese NPB schedule, due to puzzling inefficiencies in the regular season schedule that we believed could be improved. We developed a multi-round generalization of the TTP (Hoshino & Kawarabayashi, 2011c) based on Dijkstra's shortest path algorithm to create a distance-optimal intra-league schedule that reduced the total travel distance by over 60000 kilometres as compared to the 2010 NPB schedule. We elaborated further on the intricacies of intra-league scheduling in a journal paper (Hoshino & Kawarabayashi, 2011d). Inspired by the success of analyzing intra-league scheduling, we asked whether our techniques and methods could be extended to inter-league play, wondering whether the 2010 NPB schedule requiring 51134 kilometres of total team travel could be minimized to optimality. We answered that question by presenting the *Bipartite Traveling Tournament Problem* (Hoshino & Kawarabayashi, 2011b), and providing a rigorous analysis of $BTTP$ for the NPB distance matrix, producing a provably-optimal inter-league schedule requiring 42950 kilometres of total team travel (Hoshino & Kawarabayashi, 2011a).

The purpose of this paper is to expand upon our two inter-league conference papers and provide a more thorough discussion of $BTTP$ and its properties. We present a rigorous proof to a lemma we omitted due to space constraints (Hoshino & Kawarabayashi, 2011b), that is key to proving the NP-completeness of $BTTP$. We also present an application of $BTTP$ beyond Japanese baseball, by considering the problem of optimizing inter-league scheduling for the 30-team National Basketball Association (NBA) in North America. While we briefly alluded to the NBA inter-league problem (Hoshino & Kawarabayashi, 2011b), we are able to provide a full analysis in this paper.

In Section 2, we formally define $BTTP$ and discuss uniform and non-uniform schedules. In Section 3, we prove that $BTTP$ on $2n$ teams is NP-complete by obtaining a reduction from 3-SAT, the well-known NP-complete problem on boolean satisfiability (Garey & Johnson, 1979). Despite its computational intractability for general $n$, we present a simple yet powerful heuristic involving minimum-weight 4-cycle-covers and apply it to the 12-team NPB league in Japan, as well as the 30-team NBA.

In Section 4, we solve $BTTP$ for the NPB, producing an optimal schedule whose total travel distance of 42950 kilometres is 16% less than the 51134 kilometres traveled by these teams during the five weeks of inter-league play in the 2010 season. In Section 5, we produce a nearly-optimal solution to $BTTP$ for the NBA, developing a bipartite tournament schedule whose total travel distance of 537791 miles is just 3.8% higher than the trivial theoretical lower bound. In Section 6, we conclude the paper with several open problems and present directions for future research.

## 2. Definitions

Let there be $2n$ teams, with $n$ teams in each league. Let $X$ and $Y$ be the two leagues, with $X = \{x_1, x_2, \ldots, x_n\}$ and $Y = \{y_1, y_2, \ldots, y_n\}$. Let $D$ be the $2n \times 2n$ *distance matrix*, where entry $D_{p,q}$ is the distance between the home stadiums of teams $p$ and $q$. By definition, $D_{p,q} = D_{q,p}$ for all $p, q \in X \cup Y$, and all diagonal entries $D_{p,p}$ are zero. Similar to the original TTP, we require a compact double round-robin bipartite tournament schedule satisfying the following conditions:

(a) *at-most-three*: No team may have a home stand or road trip lasting more than three games.

(b) *no-repeat*: A team cannot play against the same opponent in two consecutive games.

(c) *each-venue*: For all $1 \leq i, j \leq n$, teams $x_i$ and $y_j$ play twice, once in each other's home venue.

|       | 1     | 2     | 3     | 4     | 5     | 6     |       | 1     | 2     | 3     | 4     | 5     | 6     |
|-------|-------|-------|-------|-------|-------|-------|-------|-------|-------|-------|-------|-------|-------|
| $x_1$ | $y_1$ | $y_2$ | $y_3$ | **$y_1$** | **$y_2$** | **$y_3$** | $x_1$ | $y_3$ | $y_2$ | **$y_1$** | $y_3$ | $y_1$ | **$y_2$** |
| $x_2$ | $y_2$ | $y_3$ | $y_1$ | **$y_2$** | **$y_3$** | **$y_1$** | $x_2$ | **$y_1$** | $y_3$ | **$y_2$** | $y_1$ | $y_2$ | **$y_3$** |
| $x_3$ | $y_3$ | $y_1$ | $y_2$ | **$y_3$** | **$y_1$** | **$y_2$** | $x_3$ | **$y_2$** | $y_1$ | **$y_3$** | $y_2$ | $y_3$ | **$y_1$** |
| $y_1$ | **$x_1$** | **$x_3$** | **$x_2$** | $x_1$ | $x_3$ | $x_2$ | $y_1$ | $x_2$ | **$x_3$** | $x_1$ | **$x_2$** | **$x_1$** | $x_3$ |
| $y_2$ | **$x_2$** | **$x_1$** | **$x_3$** | $x_2$ | $x_1$ | $x_3$ | $y_2$ | $x_3$ | **$x_1$** | $x_2$ | **$x_3$** | **$x_2$** | $x_1$ |
| $y_3$ | **$x_3$** | **$x_2$** | **$x_1$** | $x_3$ | $x_2$ | $x_1$ | $y_3$ | **$x_1$** | **$x_2$** | $x_3$ | $x_1$ | **$x_3$** | $x_2$ |

Table 1: Two feasible inter-league tournaments for $n = 3$.

To illustrate, Table 1 provides two feasible tournaments satisfying all of the above conditions for the case $n = 3$. In this table, as in all other schedules that will be subsequently presented, home games are marked in bold.

Following the convention of the TTP, whenever a team is scheduled for a road trip consisting of multiple away games, the team doesn't return to their home city but rather proceeds directly to their next away venue. Furthermore, we assume that every team begins the tournament at home, and returns home after playing their last away game. For example, in Table 1, team $x_1$ would travel a distance of $D_{x_1,y_1} + D_{y_1,y_2} + D_{y_2,y_3} + D_{y_3,x_1}$ when playing the schedule on the left and a distance of $D_{x_1,y_3} + D_{y_3,y_2} + D_{y_2,x_1} + D_{x_1,y_1} + D_{y_1,x_1}$ when playing the schedule on the right. The desired solution to *BTTP* is the tournament schedule that minimizes the total distance traveled by all $2n$ teams subject to the given conditions.

Define a *trip* to be a pair of consecutive games not occurring in the same city, i.e., any situation where that team doesn't play at the same location in time slots $s$ and $s+1$, and therefore has to travel from one venue to another. In Table 1, the schedule on the left has 24 total trips, while the schedule on the right has 32 trips. One may conjecture that the total distance of schedule $S_1$ is lower than the total distance of schedule $S_2$ iff $S_1$ has fewer trips than $S_2$.

To see that this is actually not the case, let the teams $x_1$, $x_3$, $y_1$, and $y_2$ be located at $(0,0)$ and let $x_2$ and $y_3$ be located at $(1,0)$. Then the schedule on the left has total distance 16 and the schedule on the right has total distance 12. So minimizing trips does not correlate to minimizing total travel distance; while the former is a trivial problem, the latter is extremely difficult, even for the case $n = 3$.

The six teams $x_1, x_2, x_3, y_1, y_2, y_3$ can be located in the Cartesian plane so that the distance-optimal solution occurs via a schedule with 27 trips, although in the majority of cases, the distance-optimal schedule consists of 24 trips, the fewest number possible. This inspires several interesting open problems which we will present at the conclusion of this paper. To provide an example with 27 trips, locate the six teams at $x_1 = (8, 0)$, $x_2 = (9, 0)$, $x_3 = (0, 4)$, $y_1 = (6, 1)$, $y_2 = (0, 7)$, and $y_3 = (3, 5)$. Then a computer search proves that the minimal distance is

$$18 + 16\sqrt{5} + 16\sqrt{2} + 3\sqrt{13} + 5\sqrt{10} + 2\sqrt{130} + \sqrt{61} \sim 133.646,$$

with equality iff the inter-league schedule is one of the two appearing in Table 2. Note that each of these 27-trip distance-optimal schedules is a mirror image of the other.

|       | 1     | 2     | 3     | 4     | 5     | 6     |     |       | 1     | 2     | 3     | 4     | 5     | 6     |
|-------|-------|-------|-------|-------|-------|-------|-----|-------|-------|-------|-------|-------|-------|-------|
| $x_1$ | $y_1$ | $y_2$ | $y_3$ | **$y_1$** | **$y_2$** | **$y_3$** | | $x_1$ | **$y_3$** | **$y_2$** | **$y_1$** | $y_3$ | $y_2$ | $y_1$ |
| $x_2$ | $y_2$ | $y_3$ | $y_1$ | **$y_2$** | **$y_3$** | **$y_1$** | | $x_2$ | **$y_1$** | **$y_3$** | **$y_2$** | $y_1$ | $y_3$ | $y_2$ |
| $x_3$ | **$y_3$** | **$y_1$** | $y_2$ | $y_3$ | $y_1$ | **$y_2$** | | $x_3$ | **$y_2$** | $y_1$ | $y_3$ | $y_2$ | **$y_1$** | **$y_3$** |
| $y_1$ | **$x_1$** | $x_3$ | **$x_2$** | $x_1$ | **$x_3$** | $x_2$ | | $y_1$ | $x_2$ | **$x_3$** | $x_1$ | **$x_2$** | $x_3$ | **$x_1$** |
| $y_2$ | **$x_2$** | **$x_1$** | **$x_3$** | $x_2$ | $x_1$ | $x_3$ | | $y_2$ | $x_3$ | $x_1$ | $x_2$ | **$x_3$** | **$x_1$** | **$x_2$** |
| $y_3$ | $x_3$ | **$x_2$** | **$x_1$** | **$x_3$** | $x_2$ | $x_1$ | | $y_3$ | $x_1$ | $x_2$ | **$x_3$** | **$x_1$** | **$x_2$** | $x_3$ |

Table 2: The 27-trip distance-optimal schedules for a special selection of 6 points.

Let *BTTP\** be the restriction of *BTTP* to the set of tournament schedules where in any given time slot, the teams in each league either all play at home, or all play on the road. For example, the left schedule in Table 1 is a feasible solution of both *BTTP* and *BTTP\**. We say that such schedules are *uniform*. While this uniformity constraint significantly reduces the number of potential tournaments, it allows us to quickly generate an approximate solution to *BTTP* from an algorithm based on minimum-weight 4-cycle-covers.

We now prove that both *BTTP* and *BTTP\** are NP-complete by obtaining a reduction from 3-SAT, the well-known NP-complete problem of deciding whether a boolean formula in conjunctive normal form with three literals per clause admits a satisfying assignment (Garey & Johnson, 1979).

## 3. Theoretical Results

To establish our reduction, we first express *BTTP* in its decision form:

   INSTANCE:

(a) $2n$ teams, in which $n$ teams belong to league $X$ and $n$ teams belong to league $Y$.

(b) A $2n \times 2n$ distance matrix $D$ whose entries are the distances between each pair of teams in $X \cup Y$.

(c) An integer $T \geq 0$.

   QUESTION: does there exist a double round-robin bipartite tournament for which:

(a) The *at-most-three*, *no-repeat*, and *each-venue* conditions are all satisfied.

(b) The sum of the distances traveled by the $2n$ teams is at most $T$.

Similarly, we can express $BTTP^*$ in its decision form, by adding the uniformity constraint (i.e., for any given time slot, a team plays at home iff every other team in that league also plays at home). We now reduce these two problems to 3-SAT.

Let $S = C_1 \wedge C_2 \wedge \ldots \wedge C_m$ be the conjunction of $m$ clauses with three literals on the variables $\{u_1, u_2, \ldots, u_l\}$. From $S$, we will define the sets $X_S$ and $Y_S$ representing the teams in leagues $X$ and $Y$. From this set of $|X_S| + |Y_S|$ vertices, we will describe a polynomial-time algorithm that constructs a complete graph and assigns edge weights to produce the distance matrix $D_S$. We then prove the existence of an integer $T = T(m)$ for which the solutions to $BTTP$ and $BTTP^*$ have total travel distance $\leq T$ iff $S$ is satisfiable. This will establish the desired polynomial-time reductions.

We can assume that literals $u_i$ and $\overline{u}_i$ occur equally often in $S$ for each $1 \leq i \leq l$. To see why, assume without loss that $u_i$ occurs less frequently than $\overline{u}_i$. By repeated addition of the tautological clause $(u_i \vee u_{i+1} \vee \overline{u}_{i+1})$, which does not affect the satisfiability of $S$, we can ensure that the number of occurrences of $u_i$ in $S$ matches that of $\overline{u}_i$.

Let $r(i)$ denote the number of occurrences of $u_i$ in $S$. In Figure 1, we present a "gadget" for each variable $u_i$, where the vertices $u_{i,r}$ and $\overline{u}_{i,r}$ correspond respectively to the $r^{\text{th}}$ occurrence of $u_i$ and $\overline{u}_i$ in $S$, vertex $a_{i,r}$ is adjacent to $\overline{u}_{i,r-1}$ and $u_{i,r}$, and vertex $b_{i,r}$ is adjacent to $u_{i,r}$ and $\overline{u}_{i,r}$. (Note: $\overline{u}_{i,0} := \overline{u}_{i,r(i)}$ for all $i$.)

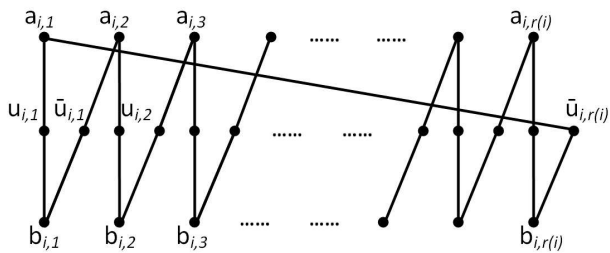

Figure 1: Gadget for 3-SAT reduction.

This gadget was used to establish the NP-completeness of deciding whether an undirected graph contains a given number of vertex-disjoint $s$-$t$ paths of a specified length (Itai, Perl, & Shiloach, 1982) and to prove that the original TTP is NP-complete (Thielen & Westphal, 2010).

There are $l$ gadgets, one for each $u_i$, $i = 1, 2, \ldots, l$. Now we define the *gadget graph* $G_S$. We create vertices $c_j$ and $d_j$ for $1 \leq j \leq m$, one pair for each clause in $S$. Join $c_j$ to $d_j$. Now connect $c_j$ to vertex $u_{i,r}$ iff clause $C_j$ contains the $r^{\text{th}}$ occurrence of $u_i$ in $S$. Similarly, connect $c_j$ to vertex $\overline{u}_{i,r}$ iff clause $C_j$ contains the $r^{\text{th}}$ occurrence of $\overline{u}_i$ in $S$.

To illustrate, let $S = C_1 \wedge C_2 \wedge C_3 \wedge C_4 \wedge C_5 \wedge C_6 \wedge C_7 \wedge C_8$, where $C_1 = (u_1 \vee u_2 \vee u_3)$, $C_2 = (\overline{u}_1 \vee \overline{u}_2 \vee \overline{u}_3)$, $C_3 = (u_1 \vee \overline{u}_2 \vee u_4)$, $C_4 = (u_2 \vee \overline{u}_3 \vee u_4)$, $C_5 = (\overline{u}_1 \vee u_3 \vee u_4)$, $C_6 = (u_1 \vee \overline{u}_2 \vee \overline{u}_4)$, $C_7 = (u_2 \vee \overline{u}_3 \vee \overline{u}_4)$, and $C_8 = (\overline{u}_1 \vee u_3 \vee \overline{u}_4)$. By definition, $S$ is an instance of 3-SAT. The gadget graph $G_S$ is given in Figure 2.

Since each literal occurs as often as its negation, and each clause has three literals, the number of clauses in $S$ must be even. Hence, $m = 2k$ for some integer $k \geq 1$. From the instance $S$, we will define a set $X_S$ with $18k$ vertices corresponding to the teams in league $X$. We will then define another set $Y_S$, with just 3 vertices (labelled $p$, $q$, and $r$), and place $6k$ teams at each of these three vertices. This will create a $36k$-team league, with $18k$ teams in both $X$ and $Y$. The weight of each edge will just correspond to the distance between the teams located at those vertices. Using the gadget graph $G_S$, we will define the edge weights in such a way that $S$ is satisfiable iff the solutions to $BTTP$ and $BTTP^*$ have total distance at most $T = T(k) = 96k^2(2900k^2 + 375k + 11)$. This will establish the desired polynomial-time reductions from 3-SAT.

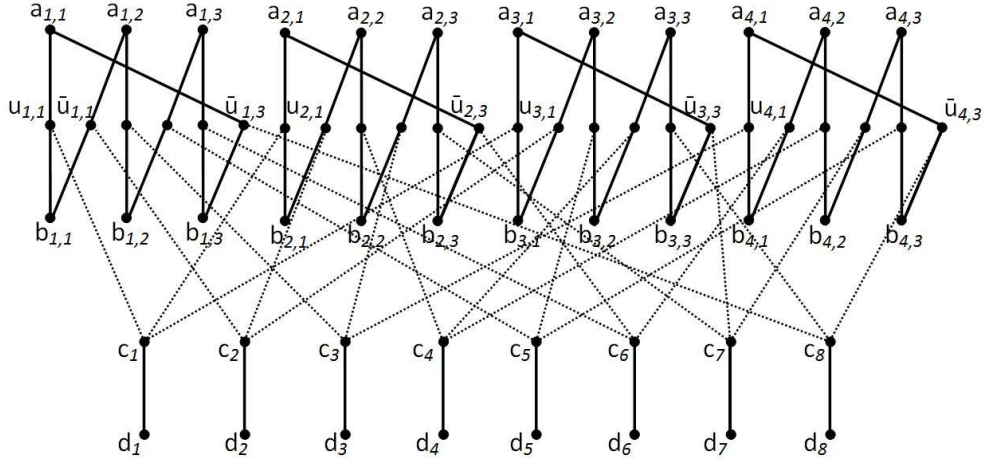

Figure 2: The gadget graph $G_S$ for the instance $S$.

We first define $X_S$. Let $C = \{c_1, c_2, \ldots, c_{2k}\}$ and $D = \{d_1, d_2, \ldots, d_{2k}\}$, which are the same set of vertices from the corresponding gadget graph $G_S$. Let $U$ be the set of $6k$ vertices of the form $u_{i,r}$ or $\overline{u}_{i,r}$ that appear in $G_S$, and let $A$ and $B$ be respectively the set of vertices of the form $a_{i,r}$ and $b_{i,r}$ that appear in $G_S$. Finally, we present two additional sets, $E = \{e_1, e_2, \ldots, e_k\}$ and $F = \{f_1, f_2, \ldots, f_k\}$, which will be matched up to the vertices of $U$ in our cycle cover.

We define $X_S = A \cup B \cup C \cup D \cup E \cup F \cup U$. Hence, $|X_S| = |A| + |B| + |C| + |D| + |E| + |F| + |U| = 3k + 3k + 2k + 2k + k + k + 6k = 18k$.

Having defined $X_S$, we now define the edge weights connecting each pair of vertices in $X_S$, thus producing a complete graph on $18k$ vertices. The weight of each edge will be a function of $k$. For readability, we will express each weight as a function of $z$, where $z := 20k + 1$. To each edge in this complete graph, we assign a weight from the set $\{z^2, z^2 + z, 2z^2 - 1\}$ as follows:

(1) A weight of $z^2$ is given to every edge that appears in the gadget graph $G_S$, the $6k^2$ edges from $U$ to $E$, and the $k$ edges connecting $e_i$ to $f_i$ (for each $1 \leq i \leq k$).

(2) A weight of $z^2 + z$ is given to the $6k^2$ edges from $U$ to $F$, the $6k$ edges connecting $A$ to $B$ through a common neighbour in $U$, and the $6k$ edges connecting $D$ to $U$ through a common neighbour in $C$.

(3) A weight of $2z^2 - 1$ is given to every other edge.

We now create an inter-league tournament with $36k$ total teams. First, we assign the $18k$ teams in league $X$ to the $18k$ vertices of graph $X_S$, where the distance between the home venues of two teams is the edge weight between the corresponding two vertices in the complete graph.

Let $Y_S = \{p, q, r\}$. Now define the $18k$ teams in league $Y$ as follows: place $6k$ teams at point $p$, $6k$ teams at point $q$, and $6k$ teams at point $r$.

Therefore, $|X_S \cup Y_S| = 18k + 3$. We now extend our complete graph on $18k$ vertices to include these three additional vertices. To assign an edge weight connecting each pair of "inter-league" vertices, we read off the matrix given in Table 3.

|  | $p \in Y_S$ | $q \in Y_S$ | $r \in Y_S$ |
|---|---|---|---|
| $a \in A$ | $z^2$ | $z^2 + z$ | $2z^2 - 1$ |
| $b \in B$ | $z^2$ | $2z^2 - 1$ | $z^2 + z$ |
| $c \in C$ | $2z^2 - 1$ | $z^2$ | $z^2 + z$ |
| $d \in D$ | $z^2$ | $2z^2 - 1$ | $z^2$ |
| $e \in E$ | $2z^2 - 1$ | $z^2 + z$ | $z^2$ |
| $f \in F$ | $z^2$ | $z^2$ | $2z^2 - 1$ |
| $u \in U$ | $z^2 + z$ | $z^2 + 2z$ | $z^2 + 2z$ |

Table 3: Weights of all edges connecting $X_S$ to $Y_S$.

For example, the edge from $c_i$ to $p$ is given a weight of $2z^2 - 1$, for all $i = 1, 2, \ldots, 2k$. We repeat the same process for each of the $7 \times 3 = 21$ pairs connecting a vertex in $X_S = A \cup B \cup C \cup D \cup E \cup F \cup U$ to a vertex in $Y_S = \{p, q, r\}$.

Finally, let the weights of edges $pq$, $pr$, and $qr$ all be $2z^2 - 1$. As a result, we have now created a complete graph on the vertex set $X_S \cup Y_S$, and assigned a weight to each edge. Moreover, the weight of each edge appears in the set $\{z^2, z^2 + z, z^2 + 2z, 2z^2 - 1\}$, where $z = 20k + 1$. As most versions of the TTP require the teams to be located at points satisfying the Triangle Inequality, we have chosen the weights in our inter-league variant BTTP to ensure that the Triangle Inequality holds for any triplet of points in $X_S \cup Y_S$.

We now partition the $18k$ vertices of $X_S$ into groups of cardinality at most three and attach them to each $y \in \{p, q, r\} = Y_S$ to produce a union of cycles of length at most 4. More formally, we define the following:

**Definition 1.** *For each $y \in Y_S$, a $y$-rooted 4-cycle-cover is a union of cycles of length at most 4, where every cycle contains $y$, no cycle contains a vertex from $Y_S \setminus \{y\}$, and every vertex of $X_S$ appears in exactly one cycle.*

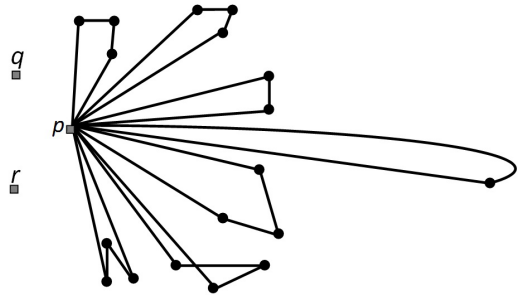

Figure 3: A $p$-rooted 4-cycle-cover with 18 vertices in set $X_S$.

To illustrate, Figure 3 gives a $p$-rooted 4-cycle-cover with $|X_S| = 18$. This definition is motivated by our tournament construction, where we will show that the total travel distance is minimized by creating a uniform schedule where each team takes the maximum number of three-game road trips to play their $18k$ away games. In the case of the $6k$ teams of $Y_S$ located at vertex $p$, their $6k$ three-game road trips will correspond to the $6k$ 4-cycles in the minimum weight $p$-rooted 4-cycle-cover. For example, if $p$-$u_{1,1}$-$c_5$-$d_5$-$p$ appears as one of the $6k$ cycles, then each team in $Y_S$ located at vertex $p$ will play three consecutive road games during the tournament against the teams of $X_S$ located at $u_{1,1}$, $c_5$, and $d_5$.

So the total distance traveled by each team at $y \in Y_S$ is bounded below by the sum of the edge weights of the minimum weight $y$-rooted 4-cycle-cover.

**Definition 2.** *We define three special types of cycles that may appear in a $p$-rooted 4-cycle-cover.*

(1) *A $(p, a, u, b, p)$-cycle is a 4-cycle with vertices $p$, $a$, $u$, $b$ in that order, where $p \in Y_S$, $a \in A$, $u \in U$, $b \in B$, where $au$ and $ub$ are both edges in the gadget graph $G_S$.*

(2) *A $(p, u, c, d, p)$-cycle is a 4-cycle with vertices $p$, $u$, $c$, $d$ in that order, where $p \in Y_S$, $u \in U$, $c \in C$, $d \in D$, where $uc$ and $cd$ are both edges in the gadget graph $G_S$.*

(3) *A $(p, u, e, f, p)$-cycle is a 4-cycle with vertices $p$, $u$, $e$, $f$ in that order, where $p \in Y_S$, $u \in U$, $e \in E$, $f \in F$, where $e$ and $f$ have the same index (i.e., $e_i$ and $f_i$ for some $1 \leq i \leq k$.)*

For example, for our instance $S$ whose gadget graph was illustrated in Figure 2, $p$-$a_{1,2}$-$\overline{u}_{1,1}$-$b_{1,1}$-$p$ is a $(p, a, u, b, p)$-cycle, but $p$-$a_{1,2}$-$\overline{u}_{1,1}$-$b_{4,2}$-$p$ is not. Similarly, $p$-$\overline{u}_{4,3}$-$c_8$-$d_8$-$p$ is a $(p, u, c, d, p)$-cycle, but $p$-$\overline{u}_{4,3}$-$c_7$-$d_7$-$p$ is not.

Following the convention of the TTP (Easton, Nemhauser, & Trick, 2002), we define $ILB_t$ to be the *individual lower bound* for team $t$. This value represents the minimum possible distance that can be traveled by team $t$ in order to complete all of their games under the constraints of $BTTP$, independent of the other teams' schedules. By definition, for each team $t$ located at $y \in Y_S$, the value of $ILB_t$ is the minimum weight of a $y$-rooted 4-cycle-cover.

Similarly, we define the *league lower bound* $LLB_T$ to be the minimum possible distance traveled by all of the teams $t$ in league $T$, and the *tournament lower bound* $TLB$ to be the minimum possible distance traveled by all the teams in both leagues. We note the following trivial inequalities:

$$TLB \geq LLB_X + LLB_Y$$
$$LLB_X \geq \sum_{t \in X} ILB_t \ , \quad LLB_Y \geq \sum_{t \in Y} ILB_t \ .$$

By definition, the solution to $BTTP$ is a tournament schedule whose total travel distance is $TLB$.

We now have all of the definitions we need to complete the proof of the NP-completeness of $BTTP$ and $BTTP^*$. We will create an inter-league tournament between the $18k$ teams of $X_S$ and the $18k$ teams of $Y_S$ (with one-third of the teams at each vertex of $Y_S$), and show that there exists a distance-optimal uniform tournament with total distance at most $T(k) = 96k^2(2900k^2 + 375k + 11)$ iff $S$ is satisfiable. This will establish our polynomial-time reduction from 3-SAT, since all of the transformations in our construction are clearly polynomial.

The desired result will follow from the next four lemmas. In each lemma, we let $K_S$ be the complete graph on the $18k + 3$ vertices of $X_S \cup Y_S$, with edge weights as described in our construction. For the interested reader, the proofs to these three lemmas appear in Appendix A.

**Lemma 1.** *The following statements are equivalent:*

(i) $S = C_1 \wedge C_2 \wedge \ldots \wedge C_{2k}$ *is satisfiable.*

(ii) *There exists a $p$-rooted 4-cycle-cover of $K_S$ with exactly $3k$ $(p, a, u, b, p)$-cycles, $2k$ $(p, u, c, d, p)$-cycles, and $k$ $(p, u, e, f, p)$-cycles.*

**Lemma 2.** *The following statements are equivalent:*

(i) *A $p$-rooted 4-cycle-cover of $K_S$ has exactly $3k$ $(p, a, u, b, p)$-cycles, $2k$ $(p, u, c, d, p)$-cycles, and $k$ $(p, u, e, f, p)$-cycles.*

(ii) *A $p$-rooted 4-cycle-cover of $K_S$ has total edge weight $k(24z^2 + 3z)$.*

**Lemma 3.** *Let $ILB_y$ be the minimum total edge weight of a $y$-rooted 4-cycle-cover of $K_S$. Then*

$$ILB_y = \begin{cases} k(24z^2 + 3z) & \text{if } y = p \\ k(24z^2 + 20z) & \text{if } y = q \\ k(24z^2 + 19z) & \text{if } y = r \end{cases}$$

Let us illustrate these three lemmas with a specific example. Let $S = C_1 \wedge C_2 \wedge C_3 \wedge C_4 \wedge C_5 \wedge C_6 \wedge C_7 \wedge C_8$ be the instance of 3-SAT whose gadget graph $G_S$ was presented in Figure 2. Recall that we defined $C_1 = (u_1 \vee u_2 \vee u_3)$, $C_2 = (\overline{u}_1 \vee \overline{u}_2 \vee \overline{u}_3)$, $C_3 = (u_1 \vee \overline{u}_2 \vee u_4)$, $C_4 = (u_2 \vee \overline{u}_3 \vee u_4)$, $C_5 = (\overline{u}_1 \vee u_3 \vee u_4)$, $C_6 = (u_1 \vee \overline{u}_2 \vee \overline{u}_4)$, $C_7 = (u_2 \vee \overline{u}_3 \vee \overline{u}_4)$, and $C_8 = (\overline{u}_1 \vee u_3 \vee \overline{u}_4)$.

Suppose $S$ is satisfiable, i.e., there is a function $\phi : \{u_1, u_2, u_3, u_4\} \to \{\text{TRUE}, \text{FALSE}\}$ so that each clause $C_i$ evaluates to TRUE for all $1 \leq i \leq 8$. By symmetry, we may assume without loss that $\phi(u_4)$ is TRUE. Then from clauses $C_6$, $C_7$, and $C_8$, we see that for $1 \leq i \leq 3$, $\phi(u_i)$ must be all TRUE or all FALSE. In the former, clause $C_2$ is FALSE, and in the latter, clause $C_1$ is FALSE. Therefore, $S$ is not satisfiable.

Since $S$ is not satisfiable, by Lemma 1, there does not exist a $p$-rooted 4-cycle-cover of $K_S$ with 12 $(p, a, u, b, p)$-cycles, 8 $(p, u, c, d, p)$-cycles, and 4 $(p, u, e, f, p)$-cycles. And by Lemma 2 and Lemma 3, the minimum weight of a $p$-rooted 4-cycle-cover of $K_S$ is strictly larger than $4(24z^2 + 3z)$.

We now show that such a (non-satisfiable) instance $S$ cannot yield a graph $K_S$ forming a distance-optimal inter-league tournament, but that a satisfiable instance $S$ indeed does.

Just as we defined special 4-cycles rooted at $p$ (e.g. $(p, a, u, b, p)$-cycles), we can similarly define 4-cycles rooted at $q$ and $r$. In Lemma 3, the lower bound $ILB_q$ occurs when the $q$-rooted 4-cycle-cover consists of $3k$ $(q, u, b, a, q)$-cycles, $2k$ $(q, c, d, u, q)$-cycles, and $k$ $(q, f, u, e, q)$-cycles, with total edge weight $3k(4z^2 + 4z) + 2k(4z^2 + 3z) + k(4z^2 + 2z) = k(24z^2 + 20z)$. The lower bound $ILB_r$ occurs when the $r$-rooted 4-cycle-cover consists of $3k$ $(r, b, a, u, r)$-cycles, $2k$ $(r, d, u, c, r)$-cycles, and $k$ $(r, e, f, u, r)$-cycles, with total edge weight $3k(4z^2 + 4z) + 2k(4z^2 + 2z) + k(4z^2 + 3z) = k(24z^2 + 19z)$. We apply this information in the following lemma in constructing our distance-optimal bipartite tournament.

**Lemma 4.** *If $S$ is satisfiable, then there exists a uniform schedule (i.e., a solution to BTTP as well as BTTP\*) whose total travel distance is $\sum ILB_t = k^2(696z^2 + 408z - 48) = 96k^2(2900k^2 + 375k + 11)$.*

*Proof.* From Lemma 1, if $S$ is satisfiable, there exists a $p$-rooted 4-cycle-cover of $K_S$ with exactly $3k$ $(p, a, u, b, p)$-cycles, $2k$ $(p, u, c, d, p)$-cycles, and $k$ $(p, u, e, f, p)$-cycles. Consider such a $p$-rooted 4-cycle-cover. We now relabel the teams in $X_S$ as follows:

First let $\{x_0, x_1, x_2\}, \{x_3, x_4, x_5\}, \ldots, \{x_{9k-3}, x_{9k-2}, x_{9k-1}\}$ be the vertices of the $3k$ $(p, a, u, b, p)$-cycles, where $x_{3i} \in A$, $x_{3i+1} \in U$, and $x_{3i+2} \in B$ for all $0 \leq i \leq 3k - 1$.

Then let $\{x_{9k}, x_{9k+1}, x_{9k+2}\}, \{x_{9k+3}, x_{9k+4}, x_{9k+5}\}, \ldots, \{x_{15k-3}, x_{15k-2}, x_{15k-1}\}$ be the vertices of the $2k$ $(p, u, c, d, p)$-cycles, where $x_{3i} \in U$, $x_{3i+1} \in C$, and $x_{3i+2} \in D$ for all $3k \leq i \leq 5k - 1$.

Finally, let $\{x_{15k}, x_{15k+1}, x_{15k+2}\}, \{x_{15k+3}, x_{15k+4}, x_{15k+5}\}, \ldots, \{x_{18k-3}, x_{18k-2}, x_{18k-1}\}$ be the vertices of the $k$ $(p, u, e, f, p)$-cycles, where $x_{3i} \in U$, $x_{3i+1} \in E$, and $x_{3i+2} \in F$ for all $5k \leq i \leq 6k - 1$.

To explain our proof more clearly, we use this relabeling of the teams in $X_S$, letting each vertex be $x_i$ for some $0 \leq i \leq 18k - 1$. We also relabel the teams in $Y_S$, so that $\{p_0, p_1, \ldots, p_{6k-1}\}$ are the teams at $p$, $\{q_0, q_1, \ldots, q_{6k-1}\}$ are the teams at $q$, and $\{r_0, r_1, \ldots, r_{6k-1}\}$ are the teams at $r$.

Since every team plays two games against each of the $18k$ teams in the other league, the tournament has $36k$ time slots. We now build a double round-robin bipartite tournament where the teams in each league play their home games in the same slots (i.e., the schedule is *uniform*.) Specifically, each team in $X_S$ will play three consecutive home games followed by three consecutive road games and repeat that pattern $6k$ times. Similarly, each team in $Y_S$ will play three consecutive road games followed by three consecutive home games and repeat that pattern until the end of the tournament. Given the way we constructed the edge weights, this is the natural way to construct a distance-optimal tournament, where each team takes as few trips as possible.

In Lemma 3, we determined the value of $ILB_v$ for each $v \in Y_S = P \cup Q \cup R$. we have $ILB_{p_i} = k(24z^2 + 3z)$, $ILB_{q_i} = k(24z^2 + 20z)$, and $ILB_{r_i} = k(24z^2 + 19z)$, for all $0 \leq i \leq 6k - 1$. Therefore, $LLB_{Y_S} \geq 6k^2(24z^2 + 3z) + 6k^2(24z^2 + 20z) + 6k^2(24z^2 + 19z) = 6k^2(72z^2 + 42z)$.

We now determine the value of $ILB_t$ for each $t \in X_S = A \cup B \cup C \cup D \cup E \cup F \cup U$. Every team $t \in X_S$ plays a road game against each of the $18k$ teams in $Y_S$, with $6k$ teams located at points $p$, $q$, and $r$. Team $t$ must make at least $\frac{6k}{3} = 2k$ trips to each of $p$, $q$, and $r$, since the maximum length of a road trip is three games. Therefore, $ILB_t \geq 2k(D_{t,p} + D_{t,q} + D_{t,r})$, where $D_{t,v}$ is the distance from $t \in X_S$ to $y \in Y_S$ for all choices of $t$ and $y$. Note that equality can occur, specifically when the road trips of team $t$ are scheduled in the most efficient way, with each trip consisting of three consecutive games against three teams located at the same point.

From Table 3, we determine that $ILB_t = 2k(D_{t,p} + D_{t,q} + D_{t,r}) = 4k(4z^2 + z - 1)$ for all $t \in A \cup B \cup C \cup E$. Similarly, $ILB_t = 4k(4z^2 - 1)$ for all $t \in D \cup F$, and $ILB_t = 4k(3z^2 + 5z)$ for all $t \in U$. Thus, we have

$$\begin{aligned}
LLB_{X_S} \\
\geq\ & 4k(4z^2 + z - 1)(|A| + |B| + |C| + |E|) + 4k(4z^2 - 1)(|D| + |F|) + 4k(3z^2 + 5z)(|U|) \\
=\ & 4k(4z^2 + z - 1)(3k + 3k + 2k + k) + 4k(4z^2 - 1)(2k + k) + 4k(3z^2 + 5z)(6k) \\
=\ & 36k^2(4z^2 + z - 1) + 12k^2(4z^2 - 1) + 24k^2(3z^2 + 5z) \\
=\ & k^2(264z^2 + 156z - 48).
\end{aligned}$$

Therefore, $TLB \geq LLB_{X_S} + LLB_{Y_S} \geq \sum ILB_t = k^2(264z^2 + 156z - 48) + 6k^2(72z^2 + 42z) = k^2(696z^2 + 408z - 48)$. To complete the proof, it suffices to construct a tournament for which each team's total travel distance matches its individual lower bound. This will prove that $TLB = \sum ILB_t = k^2(696z^2 + 408z - 48)$.

For each $0 \leq i \leq 6k - 1$ and $0 \leq j \leq 6k - 1$, we determine the opponent of teams $p_i$, $q_i$, $r_i$ in time slots $6j + 1$, $6j + 2$, $6j + 3$, $6j + 4$, $6j + 5$, and $6j + 6$. In Table 4, we provide the schedule of games in slots $6j + 1$, $6j + 2$, and $6j + 3$, where the teams in $X_S$ play at home and the teams in $Y_S$ play on the road. In this table, the function $f(i, j)$ is always reduced modulo $18k$, so that $x_{18k+z} := x_z$ for all $0 \leq z \leq 18k - 1$.

| Game | $6j+1$ | $6j+2$ | $6j+3$ | Game | $6j+1$ | $6j+2$ | $6j+3$ |
|---|---|---|---|---|---|---|---|
| $p_i$ | $x_{3(i+j)+0}$ | $x_{3(i+j)+1}$ | $x_{3(i+j)+2}$ | $p_i$ | $x_{3(i+j)+0}$ | $x_{3(i+j)+1}$ | $x_{3(i+j)+2}$ |
| $q_i$ | $x_{3(i+j)+1}$ | $x_{3(i+j)+2}$ | $x_{3(i+j)+0}$ | $q_i$ | $x_{3(i+j)+2}$ | $x_{3(i+j)+0}$ | $x_{3(i+j)+1}$ |
| $r_i$ | $x_{3(i+j)+2}$ | $x_{3(i+j)+0}$ | $x_{3(i+j)+1}$ | $r_i$ | $x_{3(i+j)+1}$ | $x_{3(i+j)+2}$ | $x_{3(i+j)+0}$ |

Table 4: The left table lists the schedule of matches when $i$ and $j$ satisfy $i + j \in \{0, 1, \ldots, 5k - 1\} \pmod{6k}$, while the right table lists the schedule when $i$ and $j$ satisfy $i + j \in \{5k, 5k + 1, \ldots, 6k - 1\} \pmod{6k}$.

Fix $i$. By this construction, each team $p_i$, $q_i$, $r_i$ will play each of $\{x_0, x_1, \ldots, x_{18k-1}\}$ on the road exactly once. Now fix $j$. In time slot $6j + k$ (with $1 \leq k \leq 3$), each team in $X_S$ appears exactly once, playing a unique opponent from $Y_S$. Each team's schedule corresponds to a rooted 4-cycle-cover. By our labeling scheme, the 4-cycle-cover of each team $p_i$ consists of $3k$ $(p, a, u, b, p)$-cycles, $2k$ $(p, u, c, d, p)$-cycles and $k$ $(p, u, e, f, p)$-cycles. Similarly, the 4-cycle-cover of each team $q_i$ consists of $3k$ $(q, u, b, a, q)$-cycles, $2k$ $(q, c, d, u, q)$-cycles, and $k$ $(q, f, u, e, q)$-cycles. Finally, the 4-cycle-cover of each team $r_i$ consists of $3k$ $(r, b, a, u, r)$-cycles, $2k$ $(r, d, u, c, r)$-cycles, and $k$ $(r, e, f, u, r)$-cycles. Therefore, each team in $Y_S$ plays their $6k$ road trips so that its total travel distance is equal to the minimum weight of a 4-cycle-cover rooted at that vertex, which by definition is equal to that team's individual lower bound. Thus, we have constructed a schedule with $LLB_{Y_S} = 6k^2(72z^2 + 42z)$.

Now we construct the other half of our schedule, where the teams in $Y_S$ play at home and the teams in $X_S$ play on the road. This is a much simpler construction. For example, one way to build this half of the schedule is to match each triplet of teams in $X_S$ (e.g. $\{x_0, x_1, x_2\}$) with a triplet of teams from the same vertex in $Y_S$ (e.g. $\{p_0, p_1, p_2\}$), and have three consecutive slots of games between the two triplets all at the home venues of the teams in league $Y_S$. Repeating this process, we can ensure that each of the $6k$ triplets in $X_S$ play all $6k$ triplets of $Y_S$ via three-game road trips. Thus, this schedule satisfies $LLB_{X_S} = k^2(264z^2 + 156z - 48)$.

All that is required when putting the schedules together is to ensure the *no-repeat* rule, which is a simple matter given all of the flexibility we have in constructing this half of the tournament schedule.

Therefore, we have completed our proof. If $S$ is satisfiable, then the bipartite tournament with teams $X_S \cup Y_S$ has $TLB = \sum ILB_t = k^2(696z^2 + 408z - 48)$. Recalling that $z = 20k+1$, we conclude that $TLB = 96k^2(2900k^2 + 375k + 11)$. □

To illustrate the preceding proof, Table 5 gives a distance-optimal schedule for the case $k = 1$, with 18 teams in each league. We just present the schedule for the teams in $Y_S$ since we can immediately derive the schedule for the teams in $X_S$ from this table. As always, home games are marked in bold.

| Game | 1 | 2 | 3 | 4 | 5 | 6 | 7 | 8 | 9 | 10 | 11 | 12 | ... | ... | 31 | 32 | 33 | 34 | 35 | 36 |
|---|---|---|---|---|---|---|---|---|---|---|---|---|---|---|---|---|---|---|---|---|
| $p_0$ | $x_0$ | $x_1$ | $x_2$ | $\mathbf{x_0}$ | $\mathbf{x_2}$ | $\mathbf{x_1}$ | $x_3$ | $x_4$ | $x_5$ | $\mathbf{x_3}$ | $\mathbf{x_5}$ | $\mathbf{x_4}$ | ... | ... | $x_{15}$ | $x_{16}$ | $x_{17}$ | $\mathbf{x_9}$ | $\mathbf{x_{11}}$ | $\mathbf{x_{10}}$ |
| $q_0$ | $x_1$ | $x_2$ | $x_0$ | $\mathbf{x_6}$ | $\mathbf{x_8}$ | $\mathbf{x_7}$ | $x_4$ | $x_5$ | $x_3$ | $\mathbf{x_9}$ | $\mathbf{x_{11}}$ | $\mathbf{x_{10}}$ | ... | ... | $x_{17}$ | $x_{15}$ | $x_{16}$ | $\mathbf{x_{15}}$ | $\mathbf{x_{17}}$ | $\mathbf{x_{16}}$ |
| $r_0$ | $x_2$ | $x_0$ | $x_1$ | $\mathbf{x_{12}}$ | $\mathbf{x_{14}}$ | $\mathbf{x_{13}}$ | $x_5$ | $x_3$ | $x_4$ | $\mathbf{x_{15}}$ | $\mathbf{x_{17}}$ | $\mathbf{x_{16}}$ | ... | ... | $x_{16}$ | $x_{17}$ | $x_{15}$ | $\mathbf{x_3}$ | $\mathbf{x_5}$ | $\mathbf{x_4}$ |
| $p_1$ | $x_3$ | $x_4$ | $x_5$ | $\mathbf{x_1}$ | $\mathbf{x_0}$ | $\mathbf{x_2}$ | $x_6$ | $x_7$ | $x_8$ | $\mathbf{x_4}$ | $\mathbf{x_3}$ | $\mathbf{x_5}$ | ... | ... | $x_0$ | $x_1$ | $x_{12}$ | $\mathbf{x_{10}}$ | $\mathbf{x_9}$ | $\mathbf{x_{11}}$ |
| $q_1$ | $x_4$ | $x_5$ | $x_3$ | $\mathbf{x_7}$ | $\mathbf{x_6}$ | $\mathbf{x_8}$ | $x_7$ | $x_8$ | $x_6$ | $\mathbf{x_{10}}$ | $\mathbf{x_9}$ | $\mathbf{x_{11}}$ | ... | ... | $x_1$ | $x_2$ | $x_0$ | $\mathbf{x_{16}}$ | $\mathbf{x_{15}}$ | $\mathbf{x_{17}}$ |
| $r_1$ | $x_5$ | $x_3$ | $x_4$ | $\mathbf{x_{13}}$ | $\mathbf{x_{12}}$ | $\mathbf{x_{14}}$ | $x_8$ | $x_6$ | $x_7$ | $\mathbf{x_{16}}$ | $\mathbf{x_{15}}$ | $\mathbf{x_{17}}$ | ... | ... | $x_2$ | $x_0$ | $x_1$ | $\mathbf{x_4}$ | $\mathbf{x_3}$ | $\mathbf{x_5}$ |
| ⋮ | ⋮ | ⋮ | ⋮ | ⋮ | ⋮ | ⋮ | ⋮ | ⋮ | ⋮ | ⋮ | ⋮ | ⋮ | ⋱ | ⋱ | ⋮ | ⋮ | ⋮ | ⋮ | ⋮ | ⋮ |
| $p_5$ | $x_{15}$ | $x_{16}$ | $x_{17}$ | $\mathbf{x_5}$ | $\mathbf{x_4}$ | $\mathbf{x_3}$ | $x_0$ | $x_1$ | $x_2$ | $\mathbf{x_1}$ | $\mathbf{x_2}$ | $\mathbf{x_0}$ | ... | ... | $x_{12}$ | $x_{13}$ | $x_{14}$ | $\mathbf{x_7}$ | $\mathbf{x_8}$ | $\mathbf{x_6}$ |
| $q_5$ | $x_{17}$ | $x_{15}$ | $x_{16}$ | $\mathbf{x_{11}}$ | $\mathbf{x_{10}}$ | $\mathbf{x_9}$ | $x_1$ | $x_2$ | $x_0$ | $\mathbf{x_7}$ | $\mathbf{x_8}$ | $\mathbf{x_6}$ | ... | ... | $x_{13}$ | $x_{14}$ | $x_{12}$ | $\mathbf{x_{13}}$ | $\mathbf{x_{14}}$ | $\mathbf{x_{12}}$ |
| $r_5$ | $x_{16}$ | $x_{17}$ | $x_{15}$ | $\mathbf{x_{17}}$ | $\mathbf{x_{16}}$ | $\mathbf{x_{15}}$ | $x_2$ | $x_0$ | $x_1$ | $\mathbf{x_{13}}$ | $\mathbf{x_{14}}$ | $\mathbf{x_{12}}$ | ... | ... | $x_{14}$ | $x_{12}$ | $x_{13}$ | $\mathbf{x_1}$ | $\mathbf{x_2}$ | $\mathbf{x_0}$ |

Table 5: A distance-optimal inter-league tournament with 18 teams in each league.

Having provided all of the lemmas, we can now prove the main theorem of this paper.

**Theorem 1.** *BTTP and BTTP\* are NP-complete.*

*Proof.* Let $S$ be an instance of 3-SAT with $2k$ clauses, and create sets $X_S$ and $Y_S$, with edge weights as described in our construction. Consider an inter-league tournament between the $18k$ teams at $X_S$ and the $18k$ teams at $Y_S$ (with one-third of the teams at each vertex of $Y_S$).

By Lemma 4, if $S$ is satisfiable, then there exists a uniform double round-robin bipartite tournament with total distance at most $96k^2(2900k^2 + 375k + 11)$. By definition, this tournament is a feasible solution to $BTTP$ and $BTTP^*$. We now prove the converse.

Let $T(k) = 96k^2(2900k^2 + 375k + 11)$. Consider an inter-league tournament between these $36k$ teams with total travel distance at most $T(k)$. By Lemma 4, $T(k) = \sum ILB_t$. Hence, *every* team $t \in X_S \cup Y_S$ must travel the shortest possible distance of $ILB_t$ to play all of their games. By Lemma 3, this implies that every team located at $p \in Y_S$ must travel a distance of $ILB_p = k(24z^2 + 3z)$.

By Lemma 2, if each team $p \in Y_S$ travels a distance of $k(24z^2 + 3z)$, then the graph $K_S$ must contain exactly $3k$ $(p, a, u, b, p)$-cycles, $2k$ $(p, u, c, d, p)$-cycles, and $k$ $(p, u, e, f, p)$-cycles. And by Lemma 1, this occurs iff $S$ is satisfiable.

Therefore, we have constructed a double round-robin bipartite tournament $K_S$ on $36k$ teams with distance matrix $D_S$ for which the solutions to $BTTP$ and $BTTP^*$ have total distance $\leq T(k)$ iff the instance $S$ with $2k$ clauses is satisfiable. This establishes the desired polynomial-time reduction from 3-SAT, proving the NP-hardness of $BTTP$ and $BTTP^*$. Finally, we note that both problems are clearly in NP, since the distance traveled by the teams can be calculated in polynomial time. Therefore, we conclude that $BTTP$ and $BTTP^*$ are NP-complete. □

To illustrate the difference between $BTTP$ and $BTTP^*$, we provide a concrete illustration for the case $n = 3$. Let the teams be $X = \{x_1, x_2, x_3\}$ and $Y = \{y_1, y_2, y_3\}$. In Figure 4, the teams are located on the Cartesian plane, where $x_1$ and $x_2$ represent the same point, $y_1$ and $y_2$ represent the same point, and the non-negative distances $a$, $b$, $c$ satisfy the Pythagorean equation $a^2 + b^2 = c^2$.

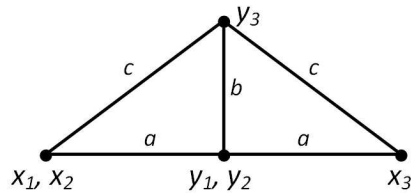

Figure 4: Illustration of $BTTP$ for the case $n = 3$.

It is straightforward to show that $ILB_{x_1} = ILB_{x_2} = ILB_{x_3} = a + b + c$, $ILB_{y_1} = ILB_{y_2} = 4a$ and $ILB_{y_3} = 2a + 2c$. Hence, $TLB \geq LLB_X + LLB_Y \geq (3a + 3b + 3c) + (10a + 2c) = 13a + 3b + 5c$.

In order for $TLB = 13a + 3b + 5c$, each of the teams in $X$ must play a three-game road stand with consecutive games against $y_1$ and $y_2$, and each of the teams in $Y$ must play a three-game road stand with consecutive games against $x_1$ and $x_2$. One can quickly show that such a scenario is impossible, but that a "nearly-best" schedule can be achieved by making either $y_1$ or $y_2$ take an extra trip, adding $2a$ to the total distance. Hence, the solution to $BTTP$ must have distance at least $15a + 3b + 5c$.

For $BTTP^*$, we have $TLB \geq 16a + 4b + 4c$ since $LLB_X = 4a + 4b + 2c$ and $LLB_Y = 12a + 2c$ in a uniform schedule. For both problems, we justify optimality by presenting a feasible tournament satisfying the stated tournament lower bounds. This is presented in Table 6.

| Team | 1 | 2 | 3 | 4 | 5 | 6 |
|---|---|---|---|---|---|---|
| $x_1$ | $y_1$ | $y_2$ | $y_3$ | $\mathbf{y_1}$ | $\mathbf{y_2}$ | $\mathbf{y_3}$ |
| $x_2$ | $y_2$ | $y_3$ | $y_1$ | $\mathbf{y_2}$ | $\mathbf{y_3}$ | $\mathbf{y_1}$ |
| $x_3$ | $y_3$ | $y_1$ | $y_2$ | $\mathbf{y_3}$ | $\mathbf{y_1}$ | $\mathbf{y_2}$ |
| $y_1$ | $\mathbf{x_1}$ | $\mathbf{x_3}$ | $\mathbf{x_2}$ | $x_1$ | $x_3$ | $x_2$ |
| $y_2$ | $\mathbf{x_2}$ | $\mathbf{x_1}$ | $\mathbf{x_3}$ | $x_2$ | $x_1$ | $x_3$ |
| $y_3$ | $\mathbf{x_3}$ | $\mathbf{x_2}$ | $\mathbf{x_1}$ | $x_3$ | $x_2$ | $x_1$ |

| Team | 1 | 2 | 3 | 4 | 5 | 6 |
|---|---|---|---|---|---|---|
| $x_1$ | $\mathbf{y_1}$ | $\mathbf{y_3}$ | $\mathbf{y_2}$ | $y_1$ | $y_3$ | $\mathbf{y_2}$ |
| $x_2$ | $\mathbf{y_2}$ | $\mathbf{y_1}$ | $\mathbf{y_3}$ | $y_2$ | $y_1$ | $y_3$ |
| $x_3$ | $\mathbf{y_3}$ | $\mathbf{y_2}$ | $\mathbf{y_1}$ | $y_3$ | $y_2$ | $y_1$ |
| $y_1$ | $x_1$ | $x_2$ | $x_3$ | $\mathbf{x_1}$ | $\mathbf{x_2}$ | $\mathbf{x_3}$ |
| $y_2$ | $x_2$ | $x_3$ | $\mathbf{x_1}$ | $\mathbf{x_2}$ | $\mathbf{x_3}$ | $x_1$ |
| $y_3$ | $x_3$ | $x_1$ | $x_2$ | $\mathbf{x_3}$ | $\mathbf{x_1}$ | $\mathbf{x_2}$ |

Table 6: Solutions to $BTTP^*$ and $BTTP$, with total distance $16a + 4b + 4c$ and $15a + 3b + 5c$, respectively.

Scheduling Bipartite Tournaments to Minimize Total Travel DistanceFor this example, the solution to $BTTP$ requires 25 trips, one more trip than the solution to $BTTP^*$, yet the tournament lower bound is reduced by $a + b - c > 0$. As we will see in the concluding section, there are many examples where the solution to the $n = 3$ $BTTP$ requires more than 24 trips.

To illustrate with an example for the case $n = 6$, consider a 12-team league with six teams in each of $X$ and $Y$. Place three points $A, B, C$ equally spaced around a unit circle, so that $\triangle ABC$ is equilateral. Place $\{x_1, x_2\}$ at $A$, $\{x_3, x_4\}$ at $B$, and $\{x_5, x_6\}$ at $C$. Now place $\{y_1, y_2, \ldots, y_6\}$ at the centre of the circle. Then the best lower bound of $ILB_{y_j} = 6$ occurs when $y_j$ plays two-game road trips against $\{x_1, x_2\}$, $\{x_3, x_4\}$, $\{x_5, x_6\}$ in pairs rather than in three-game road trips such as $\{x_1, x_2, x_3\}$ and $\{x_4, x_5, x_6\}$, which has total distance $4 + 2\sqrt{3} > 6$. And clearly the best lower bound $ILB_{x_i} = 4$ occurs when $x_i$ plays three-game road trips against the teams in $Y$, making just two trips to the centre of the circle.

Table 7 provides an distance-optimal schedule which is uniform, thus proving that for this simple example, the solution to $BTTP$ is the same as $BTTP^*$. However, note that unlike our proof of Theorem 1, in this 12-team scenario, the best schedule requires 102 total trips, six more than the fewest possible number of total trips.

|       | 1     | 2     | 3     | 4     | 5     | 6     | 7     | 8     | 9     | 10    | 11    | 12    |
|-------|-------|-------|-------|-------|-------|-------|-------|-------|-------|-------|-------|-------|
| $x_1$ | **$y_1$** | **$y_2$** | $y_1$ | $y_2$ | $y_3$ | **$y_4$** | $y_3$ | $y_4$ | $y_5$ | $y_6$ | **$y_5$** | **$y_6$** |
| $x_2$ | **$y_2$** | **$y_1$** | $y_2$ | $y_3$ | $y_1$ | **$y_3$** | $y_4$ | $y_5$ | $y_6$ | $y_4$ | **$y_6$** | **$y_5$** |
| $x_3$ | **$y_3$** | **$y_4$** | $y_3$ | $y_1$ | $y_2$ | **$y_6$** | $y_5$ | $y_6$ | $y_4$ | $y_5$ | **$y_1$** | **$y_2$** |
| $x_4$ | **$y_4$** | **$y_3$** | $y_4$ | $y_5$ | $y_6$ | **$y_5$** | $y_6$ | $y_1$ | $y_2$ | $y_3$ | **$y_2$** | **$y_1$** |
| $x_5$ | **$y_5$** | **$y_6$** | $y_5$ | $y_6$ | $y_4$ | **$y_2$** | $y_1$ | $y_2$ | $y_3$ | $y_1$ | **$y_3$** | **$y_4$** |
| $x_6$ | **$y_6$** | **$y_5$** | $y_6$ | $y_4$ | $y_5$ | **$y_1$** | $y_2$ | $y_3$ | $y_1$ | $y_2$ | **$y_4$** | **$y_3$** |
| $y_1$ | $x_1$ | $x_2$ | **$x_1$** | $x_3$ | **$x_2$** | $x_6$ | $x_5$ | **$x_4$** | **$x_6$** | **$x_5$** | $x_3$ | $x_4$ |
| $y_2$ | $x_2$ | $x_1$ | **$x_2$** | $x_1$ | **$x_3$** | $x_5$ | $x_6$ | **$x_5$** | **$x_4$** | **$x_6$** | $x_4$ | $x_3$ |
| $y_3$ | $x_3$ | $x_4$ | **$x_3$** | $x_2$ | **$x_1$** | $x_2$ | $x_1$ | **$x_6$** | **$x_5$** | **$x_4$** | $x_5$ | $x_6$ |
| $y_4$ | $x_4$ | $x_3$ | **$x_4$** | $x_6$ | **$x_5$** | $x_1$ | $x_2$ | **$x_1$** | **$x_3$** | **$x_2$** | $x_6$ | $x_5$ |
| $y_5$ | $x_5$ | $x_6$ | **$x_5$** | $x_4$ | **$x_6$** | $x_4$ | $x_3$ | **$x_2$** | **$x_1$** | **$x_3$** | $x_1$ | $x_2$ |
| $y_6$ | $x_6$ | $x_5$ | **$x_6$** | $x_5$ | **$x_4$** | $x_3$ | $x_4$ | **$x_3$** | **$x_2$** | **$x_1$** | $x_2$ | $x_1$ |

Table 7: Solution to $BTTP$ and $BTTP^*$ for a scenario with $n = 6$.

Having provided simple illustrations for $n = 3$ and $n = 6$, we now analyze $BTTP$ for two professional sports leagues, namely the Nippon Professional Baseball league (with $n = 6$) and the National Basketball Association (with $n = 15$).

## 4. Japanese Baseball

Nippon Professional Baseball (NPB) is Japan's largest professional sports league. In the NPB, the teams are split into two leagues of six teams, with each team playing 120 intra-league and 24 inter-league games during the regular season. The intra-league problem was analyzed recently by the authors (Hoshino & Kawarabayashi, 2011c), where we developed a multi-round generalization of the TTP based on Dijkstra's shortest path algorithm and applied it to produce a distance-optimal schedule reducing the total travel distance by over 60000 kilometres (a 25% reduction) as compared to the 2010 NPB intra-league schedule (Hoshino & Kawarabayashi, 2011d). Given that Japan is a small island country, a 60000 kilometre reduction represents a significant amount.

105

We now consider the inter-league problem, where the six teams in the NPB Pacific League each play four games against all six teams in the NPB Central League, with one two-game *set* played at the home of the Pacific League team, and the other two-game set played at the home of the Central League team. All inter-league games take place during a five-week stretch between mid-May and mid-June, with no intra-league games occurring during that period. Thus, the NPB inter-league scheduling problem is precisely $BTTP$, for the case $n = 6$.

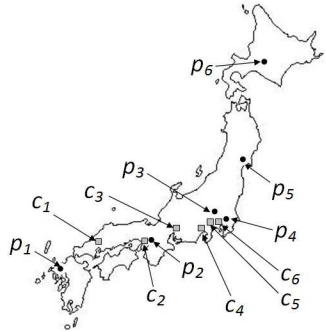

Figure 5: Location of the 12 teams in the NPB.

The locations of each team's home stadium is marked in Figure 5. For readability, we label each team as follows: the Pacific League teams are $p_1$ (Fukuoka), $p_2$ (Orix), $p_3$ (Saitama), $p_4$ (Chiba), $p_5$ (Tohoku), $p_6$ (Hokkaido), and the Central League teams are $c_1$ (Hiroshima), $c_2$ (Hanshin), $c_3$ (Chunichi), $c_4$ (Yokohama), $c_5$ (Yomiuri), and $c_6$ (Yakult). The actual $12 \times 12$ NPB distance matrix is provided in Appendix B.

We now solve $BTTP$ for the NPB, producing an inter-league schedule requiring 42950 kilometres of total travel, representing a 16% reduction compared to the 51134 kilometres traveled by these teams during the 2010 inter-league schedule (Hoshino & Kawarabayashi, 2011a). To accomplish this, we present two powerful reduction heuristics. To motivate these heuristics, we first require several key definitions.

For each $t \in X \cup Y$, let $S_t$ be the set of possible schedules that can be played by team $t$ satisfying the *at-most-three* and *each-venue* constraints. Let $\pi_t \in S_t$ be a possible schedule for team $t$. For each $\pi_t$, we just list the opponents of the six *road* sets, and ignore the home sets, since we can determine the total distance traveled by team $t$ just from the road sets. To give an example, below is a feasible schedule $\pi_{x_1} \in S_{x_1}$ for the case $n = 6$:

|       | 1     | 2     | 3 | 4 | 5     | 6     | 7     | 8 | 9 | 10 | 11    | 12 |
|-------|-------|-------|---|---|-------|-------|-------|---|---|----|-------|----|
| $x_1$ | $y_1$ | $y_6$ | *y* | *y* | $y_3$ | $y_5$ | $y_4$ | *y* | *y* | *y* | $y_2$ | *y* |

In the following team schedule $\pi_{x_1}$, each *y* represents a home set played by $x_1$ against a unique opponent in $Y$. Note that $\pi_{x_1}$ satisfies the *at-most-three* and *each-venue* constraints.

Let $\Phi = (\pi_{x_1}, \pi_{x_2}, \ldots, \pi_{x_n}, \pi_{y_1}, \pi_{y_2}, \ldots, \pi_{y_n})$, where $\pi_t \in S_t$ for each $t \in X \cup Y$. Since road sets of $X$ correspond to home sets of $Y$ and vice-versa, it suffices to list just the time slots and opponents of the $n$ road sets in each $\pi_t$, since we can then uniquely determine the full schedule of $2n$ sets for every team $t \in X \cup Y$, thus producing an inter-league tournament schedule $\Phi$. We note that $\Phi$ is a feasible solution to $BTTP$ iff each team plays a unique opponent in every time slot, and no team schedule $\pi_t$ violates the *no-repeat* constraint. In this section, we will frequently refer to *team* schedules $\pi_t$ and *tournament* schedules $\Phi$. From the context it will be clear whether the schedule is for an individual team $t \in X \cup Y$, or for all $2n$ teams in $X \cup Y$.

As before, define $ILB_t$ to be the individual lower bound of team $t$, the minimum possible distance that can be traveled by team $t$ in order to complete its $2n$ sets.

For each $\pi_t \in S_t$, let $d(\pi_t)$ be the integer for which $d(\pi_t) + ILB_t$ equals the total distance traveled by team $t$ when playing the schedule $\pi_t$. By definition, $d(\pi_t) \geq 0$.

For each $\Phi = (\pi_{x_1}, \ldots, \pi_{x_n}, \pi_{y_1}, \ldots, \pi_{y_n})$, define

$$d(\Phi) = \sum_{t \in X \cup Y} d(\pi_t).$$

Since $\sum ILB_t$ is fixed, the optimal solution to $BTTP$ is the schedule $\Phi$ for which $d(\Phi)$ is minimized. This is the motivation for the function $d(\Phi)$.

For each subset $S_t^* \subseteq S_t$, define the *lower bound* function

$$B(S_t^*) = \min_{\pi_t \in S_t^*} d(\pi_t).$$

If $S_t^* = S_t$, then $B(S_t^*) = 0$ by the definition of $ILB_t$. For each subset $S_t^*$, we define $|S_t^*|$ to be its cardinality.

For example, consider the $n = 6$ instance in Table 7, where we located the six teams in league $X$ so that two teams were assigned to each vertex of an equilateral triangle. As mentioned at the end of Section 3, $ILB_{y_1} = 6$, with equality occurring iff $y_1 \in Y$ plays two-set road trips against $\{x_1, x_2\}$, $\{x_3, x_4\}$, and $\{x_5, x_6\}$. Now let $S_{y_1}^*$ be the restriction of $S_{y_1}$ to the subset of schedules where $y_1$ starts with three consecutive road sets against teams $x_1$, $x_2$, and $x_3$. Then any such schedule must have total distance $\geq 4 + 2\sqrt{3}$, implying that $B(S_{y_1}^*) = 4 + 2\sqrt{3} - ILB_{y_1} = 2\sqrt{3} - 2 > 0$.

If $n$ is a multiple of 3, we define for each team the set $R_3^t$ as the subset of schedules in $S_t$ for which the $n$ road sets occur in $\frac{n}{3}$ blocks of three (i.e., team $t$ takes $\frac{n}{3}$ three-set road trips). For example, in Table 5 (which has $n = 18$), every team $t$ plays a schedule $\pi_t \in R_3^t$.

Finally, we define $\Gamma$ to be a *global constraint* that fixes some subset of matches, and $S_t^\Gamma$ to be the subset of schedules in $S_t$ which are consistent with that global constraint. For example, if $\Gamma$ is the simple constraint that forces $y_2$ to play against $x_1$ at home in time slot 3, then $S_{x_1}^\Gamma$ would only consist of the team schedules where slot 3 is a road set against $y_2$. If $\Gamma$ is a much more complex global constraint (e.g. where the number of fixed matches is large), then each $|S_t^\Gamma|$ will be significantly less than $|S_t|$.

To illustrate this concept, consider the above $n = 6$ instance and the global constraint $\Gamma$ that $y_1$ starts with three consecutive road sets against teams $x_1$, $x_2$, and $x_3$ (in that order). One can show that there are only 34 valid home-road patterns in $S_{y_1}$, including RRR-HHH-RRR-HHH and RRR-H-R-HH-R-HH-R-H. For each home-road pattern, there are $3! = 6$ ways to assign $\{x_4, x_5, x_6\}$ to the last three road sets. Thus, $|S_{y_1}^\Gamma| = 34 \times 3! = 204$, which is significantly less than $|S_{y_1}|$ which can be shown to equal $616 \times 6! = 443520$.

This simple notion of global constraints inspires our first result, a powerful reduction heuristic that drastically cuts down the computation time.

**Proposition 1.** *Let $M$ be a fixed positive integer. For any global constraint $\Gamma$, define for each $t \in X \cup Y$,*

$$Z_t = \left\{ \pi_t \in S_t^\Gamma : d(\pi_t) \leq M + B(S_t^\Gamma) - \sum_{u \in X \cup Y} B(S_u^\Gamma) \right\}.$$

*If $\Phi = (\pi_{x_1}, \ldots, \pi_{x_n}, \pi_{y_1}, \ldots, \pi_{y_n})$ is a feasible tournament schedule consistent with $\Gamma$ so that $d(\Phi) \leq M$, then for each $t \in X \cup Y$, team $t$'s schedule $\pi_t$ appears in $Z_t$.*

*Proof.* Consider all tournament schedules consistent with $\Gamma$. If there is no $\Phi$ with $d(\Phi) \leq M$, then there is nothing to prove. So assume some schedule $\Phi$ satisfies $d(\Phi) \leq M$. Letting $Q = \sum_{u \in X \cup Y} B(S_u^\Gamma)$, we have $M \geq d(\Phi) = \sum_{u \in X \cup Y} d(\pi_u) \geq \sum_{u \in X \cup Y} B(S_u^\Gamma)$, so that $M \geq Q$.

If $\pi_t \in Z_t$, then $Z_t \subseteq S_t^\Gamma$ implying that $d(\pi_t) \geq B(S_t^\Gamma)$. Now suppose there exists some $v \in X \cup Y$ with $\pi_v \notin Z_v$. Since $\pi_v$ is consistent with $\Gamma$, $\pi_v \in S_v^\Gamma$ and $d(\pi_v) > M + B(S_v^\Gamma) - Q \geq B(S_v^\Gamma)$. This is a contradiction, as

$$\begin{aligned} d(\Phi) &= d(\pi_v) + \sum_{u \in X \cup Y, u \neq v} d(\pi_u) \\ &> (M + B(S_v^\Gamma) - Q) + \sum_{u \in X \cup Y, u \neq v} B(S_u^\Gamma) \\ &= (M + B(S_v^\Gamma) - Q) + (Q - B(S_v^\Gamma)) = M. \end{aligned}$$

Hence, if $\Phi = (\pi_{x_1}, \ldots, \pi_{x_n}, \pi_{y_1}, \ldots, \pi_{y_n})$ is a feasible tournament schedule consistent with $\Gamma$ so that $d(\Phi) \leq M$, then $\pi_t \in Z_t$ for all $t \in X \cup Y$. □

Proposition 1 shows how to perform some reduction *prior* to propagation, and may be applicable to other problems. To apply this proposition, we will reduce $BTTP$ to $k$ scenarios where in each scenario the six home sets for four of the Pacific League teams are pre-determined. Expressing these scenarios as the global constraints $\Gamma_1, \Gamma_2, \ldots, \Gamma_k$, each $\Gamma_i$ fixes 24 of the 72 total matches.

For every $\Gamma_i$, we determine $Z_{c_j}$ for the Central League teams and by setting a low threshold $M$, we show that each $|Z_{c_j}|$ is considerably smaller than $|S_{c_j}^\Gamma|$, thus reducing the search space to an amount that can be quickly analyzed. From there, we run a simple six-loop that generates all 6-tuples $(\pi_{c_1}, \pi_{c_2}, \pi_{c_3}, \pi_{c_4}, \pi_{c_5}, \pi_{c_6})$ that can appear in a feasible schedule $\Phi$ with $d(\Phi) \leq M$. By Proposition 1, each $\pi_{c_j} \in Z_{c_j}$ for $1 \leq j \leq 6$. From this list of possible 6-tuples, we can quickly find the optimal schedule $\Phi$ which corresponds to the solution to $BTTP$.

We now present a result that works only for the case $n = 6$, when two teams from one league are located quite far from the other 10 teams, forcing the distance-optimal schedule $\Phi$ to have a particular structure.

**Proposition 2.** *Let $M$ be a fixed positive integer, and define $S_t^* = \{\pi_t \in S_t : d(\pi_t) \leq M\}$. Suppose there exist two teams $x_i, x_j \in X = \{x_1, x_2, \ldots, x_6\}$ for which $S_{x_i}^* \subseteq R_3^{x_i}$, $S_{x_j}^* \subseteq R_3^{x_j}$, and for each team $y_k \in Y$, every schedule in $S_{y_k}^*$ has the property that $y_k$ plays their road sets against $x_i$ and $x_j$ in two consecutive time slots. If $\Phi = (\pi_{x_1}, \ldots, \pi_{x_6}, \pi_{y_1}, \ldots, \pi_{y_6})$ is a feasible tournament schedule with $d(\Phi) \leq M$ where each $\pi_t \in S_t^*$, then the team schedules* $\pi_{x_i}$ and $\pi_{x_j}$ both have the home-road pattern HH-RRR-HH-RRR-HH; moreover, each team's six home slots must have the following structure for some permutation $(a,b,c,d,e,f)$ of $\{1,2,3,4,5,6\}$:

|       | 1    | 2    | 3 | 4 | 5 | 6    | 7    | 8 | 9 | 10 | 11   | 12   |
|-------|------|------|---|---|---|------|------|---|---|----|------|------|
| $x_i$ | $y_a$| $y_b$| $y$ | $y$ | $y$ | $y_c$| $y_d$| $y$ | $y$ | $y$ | $y_e$| $y_f$|
| $x_j$ | $y_b$| $y_a$| $y$ | $y$ | $y$ | $y_d$| $y_c$| $y$ | $y$ | $y$ | $y_f$| $y_e$|

*Proof.* We first note that if $\pi_{x_i}$ and $\pi_{x_j}$ have the above structure, they satisfy all the given conditions since $\pi_{x_i} \in R_3^{x_i}$, $\pi_{x_j} \in R_3^{x_j}$, and every team $y_k \in Y$ plays road sets against $x_i$ and $x_j$ in two consecutive time slots. For example, $y_d$ plays road sets against $x_j$ in slot 6 and against $x_i$ in slot 7. We now prove that $\pi_{x_i}$ and $\pi_{x_j}$ must have this structure.

For each team $x_t \in X$ and time slot $s \in [1,12]$, define $O(x_t, s)$ to be the *opponent* of team $x_t$ in set $s$. We define $O(x_t, s)$ only when $x_t$ is playing at *home*; for the sets when $x_t$ plays on the road, $O(x_t, s)$ is undefined.

Since $\pi_{x_i} \in S_{x_i}^*$ and $S_{x_i}^* \subseteq R_3^{x_i}$, there are four possible cases to consider:

(1) $x_i$ plays set 1 at home, and sets 2 to 4 on the road.

(2) $x_i$ plays sets 1 and 2 at home, and sets 3 to 5 on the road.

(3) $x_i$ plays sets 1 to 3 at home, and sets 4 to 6 on the road.

(4) $x_i$ plays sets 1 to 3 on the road, and set 4 at home.

We examine the cases one by one. In each, suppose there exists a feasible schedule $\Phi$ satisfying all the given conditions. We finish with case (2).

In (1), let $O(x_i, 1) = y_a$. Then $O(x_j, 2) = y_a$, since $y_a$ must play road sets against $x_i$ and $x_j$ in consecutive time slots. Since $\pi_{x_j} \in R_3^{x_j}$ and $x_j$ plays at home in set 2, $x_j$ must also play at home in set 1. Thus, $O(x_j, 1) = y_b$ for some $y_b$, which forces $O(x_i, 2) = y_b$. This is a contradiction as $x_i$ plays set 2 on the road.

In (3), let $O(x_i, 1) = y_a$, $O(x_i, 2) = y_b$, and $O(x_i, 3) = y_c$. Then $O(x_j, 2) = y_a$ and $O(x_j, 4) = y_c$. Either $O(x_j, 1) = y_b$ or $O(x_j, 3) = y_b$. In either case, we violate the *at-most-three* constraint or the condition that $\pi_{x_j} \in R_3^{x_j}$.

In (4), team $x_i$ starts with a three-set road trip. In order to satisfy the *at-most-three* constraint, $\pi_{x_i}$ must have the pattern RRR-HHH-RRR-HHH. Then this reduces to case (3), as we can read the schedule $\Phi$ backwards, letting $O(x_i, 12) = y_a$, $O(x_i, 11) = y_b$, $O(x_i, 10) = y_c$, and applying the argument in the previous paragraph.

In (2), let $O(x_i, 1) = y_a$ and $O(x_i, 2) = y_b$. Then $O(x_j, 2) = y_a$ and $O(x_j, 1) = y_b$. If $O(x_j, 3) = y_c$ for some $y_c$, then $O(x_i, 4) = y_c$, forcing $x_i$ to play a single road set in slot 3. Thus, $x_j$ must play on the road in set 3, and therefore also in sets 4 and 5. Hence, both $x_i$ and $x_j$ start with two home sets followed by three road sets. Since this is the only case remaining, by symmetry $x_i$ and $x_j$ must end with two home sets preceded by three road sets. Thus, these two teams must have the pattern HH-RRR-HH-RRR-HH.

In order for each $y_k$ to play their road sets against $x_i$ and $x_j$ in two consecutive time slots, we must have $O(x_i, 6) = O(x_j, 7)$, $O(x_i, 7) = O(x_j, 6)$, $O(x_i, 11) = O(x_j, 12)$, and $O(x_i, 12) = O(x_j, 11)$. This completes the proof. □

We will use Proposition 2 to solve $BTTP$, since teams $p_5$ and $p_6$ are located quite far from the other ten teams (see Figure 5). This heuristic of isolating two teams and finding its common structure significantly reduces the search space and enables us to solve $BTTP$ for the 12-team NPB in hours rather than weeks.

By applying these results, we do not require weeks of computation time on multiple processors. With these two heuristics, $BTTP$ can be solved in less than ten hours on a single laptop. All of the code was written in Maple and compiled using Maplesoft 13 using a single Toshiba laptop under Windows with a single 2.10 GHz processor and 2.75 GB RAM.

Table 8 presents an inter-league tournament schedule $\Phi$ that is a solution to $BTTP$ with $d(\Phi) = (0 + 4 + 0 + 0 + 1 + 1) + (51 + 9 + 31 + 58 + 19 + 13) = 187$.

|       | 1     | 2     | 3     | 4     | 5     | 6     | 7     | 8     | 9     | 10    | 11    | 12    |
|-------|-------|-------|-------|-------|-------|-------|-------|-------|-------|-------|-------|-------|
| $p_1$ | $\boldsymbol{c_3}$ | $\boldsymbol{c_5}$ | $\boldsymbol{c_1}$ | $c_3$ | $c_2$ | $c_1$ | $\boldsymbol{c_6}$ | $\boldsymbol{c_2}$ | $c_4$ | $c_5$ | $c_6$ | $\boldsymbol{c_4}$ |
| $p_2$ | $\boldsymbol{c_5}$ | $\boldsymbol{c_3}$ | $c_2$ | $c_1$ | $c_3$ | $\boldsymbol{c_6}$ | $c_1$ | $c_4$ | $c_5$ | $c_6$ | $\boldsymbol{c_4}$ | $\boldsymbol{c_2}$ |
| $p_3$ | $\boldsymbol{c_4}$ | $\boldsymbol{c_2}$ | $c_6$ | $c_5$ | $c_4$ | $\boldsymbol{c_3}$ | $\boldsymbol{c_5}$ | $\boldsymbol{c_1}$ | $c_3$ | $c_2$ | $c_1$ | $\boldsymbol{c_6}$ |
| $p_4$ | $\boldsymbol{c_2}$ | $\boldsymbol{c_4}$ | $\boldsymbol{c_5}$ | $c_4$ | $c_6$ | $c_5$ | $\boldsymbol{c_3}$ | $\boldsymbol{c_6}$ | $\boldsymbol{c_1}$ | $c_3$ | $c_2$ | $c_1$ |
| $p_5$ | $\boldsymbol{c_1}$ | $\boldsymbol{c_6}$ | $c_4$ | $c_6$ | $c_5$ | $\boldsymbol{c_2}$ | $\boldsymbol{c_4}$ | $c_3$ | $c_2$ | $c_1$ | $\boldsymbol{c_5}$ | $\boldsymbol{c_3}$ |
| $p_6$ | $\boldsymbol{c_6}$ | $\boldsymbol{c_1}$ | $c_3$ | $c_2$ | $c_1$ | $\boldsymbol{c_4}$ | $\boldsymbol{c_2}$ | $c_5$ | $c_6$ | $c_4$ | $\boldsymbol{c_3}$ | $\boldsymbol{c_5}$ |
| $c_1$ | $p_5$ | $p_6$ | $p_1$ | $\boldsymbol{p_2}$ | $\boldsymbol{p_6}$ | $\boldsymbol{p_1}$ | $p_2$ | $p_3$ | $p_4$ | $\boldsymbol{p_5}$ | $p_3$ | $p_4$ |
| $c_2$ | $p_4$ | $p_3$ | $\boldsymbol{p_2}$ | $\boldsymbol{p_6}$ | $\boldsymbol{p_1}$ | $p_5$ | $p_6$ | $p_1$ | $\boldsymbol{p_5}$ | $p_3$ | $\boldsymbol{p_4}$ | $p_2$ |
| $c_3$ | $p_1$ | $p_2$ | $\boldsymbol{p_6}$ | $\boldsymbol{p_1}$ | $\boldsymbol{p_2}$ | $p_3$ | $p_4$ | $\boldsymbol{p_5}$ | $p_3$ | $\boldsymbol{p_4}$ | $p_6$ | $p_5$ |
| $c_4$ | $p_3$ | $p_4$ | $\boldsymbol{p_5}$ | $\boldsymbol{p_4}$ | $\boldsymbol{p_3}$ | $p_6$ | $p_5$ | $\boldsymbol{p_2}$ | $\boldsymbol{p_1}$ | $\boldsymbol{p_6}$ | $p_2$ | $p_1$ |
| $c_5$ | $p_2$ | $p_1$ | $p_4$ | $\boldsymbol{p_3}$ | $\boldsymbol{p_5}$ | $\boldsymbol{p_4}$ | $p_3$ | $\boldsymbol{p_6}$ | $\boldsymbol{p_2}$ | $\boldsymbol{p_1}$ | $p_5$ | $p_6$ |
| $c_6$ | $p_6$ | $p_5$ | $\boldsymbol{p_3}$ | $\boldsymbol{p_5}$ | $\boldsymbol{p_4}$ | $p_2$ | $p_1$ | $p_4$ | $\boldsymbol{p_6}$ | $\boldsymbol{p_2}$ | $\boldsymbol{p_1}$ | $p_3$ |

Table 8: Solution to $BTTP$ with total distance 42950 km.

In Table 8, we see that only seven of the twelve teams satisfy $\pi_t \in R_3^t$, namely $c_1$ and all six of the Pacific League teams. However, every Central League team in this schedule plays road sets against $p_5$ and $p_6$ in consecutive time slots. This explains why each $d(c_j)$ in $\Phi$ is small.

We claim that $\Phi$ is an optimal solution, with total distance $d(\Phi) + \sum ILB_t = 187 + 42763 = 42950$. To prove this, we set $M = 187$. Define $S_t^* = \{\pi_t \in S_t : d(\pi_t) \leq M\}$, from which we determine that $S_{p_5}^* \subseteq R_3^{p_5}$ and $S_{p_6}^* \subseteq R_3^{p_6}$.

Define $T_{c_i} \subseteq S_{c_i}^\Gamma$ to be the subset of schedules for which $c_i$ does *not* play their road sets against $p_5$ and $p_6$ in two consecutive time slots. From this, we can show that $B(T_{c_3}) = 153$, and that $B(T_{c_j}) > M = 187$ for $j \in \{1, 2, 4, 5, 6\}$. We claim that if $\Phi$ satisfies $d(\Phi) \leq 187$, then $\pi_{c_j} \notin T_{c_j}$ for all $1 \leq j \leq 6$.

It suffices to prove the claim for $j = 3$. There are 144 schedules in $T_{c_3}$, all of which belong to the set $R_3^{c_3}$. For example, one such schedule $\pi_{c_3}$ is

|       | 1   | 2     | 3     | 4     | 5   | 6   | 7   | 8     | 9     | 10    | 11  | 12  |
|-------|-----|-------|-------|-------|-----|-----|-----|-------|-------|-------|-----|-----|
| $c_3$ | $\boldsymbol{p}$ | $p_2$ | $p_1$ | $p_6$ | $\boldsymbol{p}$ | $\boldsymbol{p}$ | $\boldsymbol{p}$ | $p_3$ | $p_4$ | $p_5$ | $\boldsymbol{p}$ | $\boldsymbol{p}$ |

Suppose there exists a tournament schedule $\Phi$ with $d(\Phi) \leq 187$ and $\pi_{c_3} \in T_{c_3}$. There are nine possible home-road patterns for $\pi_{p_5} \in R_3^{p_5}$ (e.g. HHH-RRR-H-RRR-HH and H-RRR-HHH-RRR-HH), each of which gives rise to $6! = 720$ possible orderings for the six home sets. Thus, there are $9 \times 720 = 6480$ ways we can select the time slots and opponents for the six home sets in $\pi_{p_5}$. Similarly, there are 6480 ways to do this for $\pi_{p_6}$. A simple Maplesoft procedure shows that only 140 of the $6480^2$ possible pairs $(\pi_{p_5}, \pi_{p_6})$ are consistent with at least one $\pi_{c_3} \in T_{c_3}$.

For each of these 140 cases, define the global constraints $\Gamma_1, \Gamma_2, \ldots, \Gamma_{140}$, obtained from fixing the twelve home sets in $\{\pi_{p_5}, \pi_{p_6}\}$. For each $\Gamma_k$, define for each $j \in \{1, 2, 4, 5, 6\}$ the set $Z_{c_j} = \{\pi_{c_j} \in S_{c_j}^{\Gamma_k} : d(\pi_{c_j}) \leq M - B(T_{c_3}) = 34\}$. Then we run our six-loop to compute all possible 6-tuples $(\pi_{c_1}, \pi_{c_2}, \ldots, \pi_{c_6})$ satisfying the given conditions with $\pi_{c_3} \in T_{c_3}$ and $\pi_{c_j} \in Z_{c_j}$ for $j \neq 3$. Within twenty minutes, Maplesoft solves all 140 cases and returns no feasible 6-tuples that can appear in a schedule $\Phi$ with $d(\Phi) \leq 187$.

Therefore, in $\Phi$, each $c_j$ must play road sets against $p_5$ and $p_6$ in consecutive time slots. Thus, teams $p_5$ and $p_6$ satisfy the conditions of Proposition 2. Hence, the home-road pattern of $\pi_{p_5}$ and $\pi_{p_6}$ in $\Phi$ must be HH-RRR-HH-RRR-HH.

Without loss, assume that $p_5$ plays a home set against $c_1$ within the first *six* time slots; otherwise we can read the schedule backwards by symmetry. Thus, there are $\frac{6!}{2} = 360$ ways to assign opponents to the six home sets in $\pi_{p_5}$. By Proposition 2, each of these 360 arrangements uniquely determines the six home sets in $\pi_{p_6}$.

A short calculation shows that in order for $d(\Phi) \leq M = 187$, teams $p_1$ and $p_3$ must also play their six road sets in two blocks of three. In other words, $\pi_{p_1} \in R_3^{p_1}$ and $\pi_{p_3} \in R_3^{p_3}$. As mentioned earlier, there are $9 \times 6!$ possible ways to select the six home sets for each of $\pi_{p_1}$ and $\pi_{p_3}$.

Thus, there are $360 \times (9 \cdot 6!) \times (9 \cdot 6!)$ ways we can select the 24 home sets played by the teams in $\{p_1, p_3, p_5, p_6\}$. We eliminate all scenarios in which some $p_i$ and $p_j$ play against some $c_k$ in the same time slot. For the possibilities that remain, we create a global constraint to apply Proposition 1.

Let $\{\Gamma_1, \Gamma_2, \ldots, \Gamma_k\}$ be the complete set of global constraints derived from the above process, where each $\Gamma_i$ fixes 24 of the 72 matches, corresponding to the home sets of $\{p_1, p_3, p_5, p_6\}$. The reduction heuristic of Proposition 1 allows us to quickly verify the existence of feasible tournament schedules $\Phi$ consistent with $\Gamma_i$ for which $d(\Phi) \leq M$.

To explain this procedure, let us illustrate with the inter-league schedule in Table 8. Let $\Gamma$ be the constraint that fixes the 24 home sets of teams $p_1$, $p_3$, $p_5$, and $p_6$ in that table. Then $S_{c_5}^{\Gamma}$, defined as the subset of schedules in $S_{c_5}$ consistent with $\Gamma$, consists only of team schedules $\pi_{c_5}$ for which $c_5$ plays road sets against $p_1$ in slot 2, $p_3$ in slot 7, $p_5$ in slot 11, and $p_6$ in slot 12.

We find that there are only 11 schedules $\pi_{c_5} \in S_{c_5}^{\Gamma}$ with $d(\pi_{c_5}) \leq M$ that are consistent with $\Gamma$. Furthermore, each $d(\pi_{c_5}) \in \{19, 41, 46, 48\}$, implying that $B(S_{c_5}^{\Gamma}) = 19$. Similarly, we can calculate the other values of $B(S_{c_j}^{\Gamma})$.

We find that $\sum_{j=1}^{6} B(S_{p_j}^{\Gamma}) = 0$ and $\sum_{j=1}^{6} B(S_{c_j}^{\Gamma}) = 51 + 9 + 31 + 58 + 19 + 13 = 181$, implying that $Z_{c_5} = \{\pi_{c_5} \in S_{c_5}^{\Gamma} : d(\pi_{c_5}) \leq 187 + 19 - 181 = 25\}$. Hence, $Z_{c_5}$ reduces to just the *two* schedules with $d(\pi_{c_5}) = 19$, including the team schedule $\pi_{c_5}$ in Table 8.

By Proposition 1, any schedule $\Phi$ consistent with $\Gamma$ satisfying $d(\Phi) \leq M$ must have the property that $\pi_t \in Z_t$ for each team $t$. Since each $|Z_{c_j}|$ is small, the calculation is extremely fast. Of course, if any $|Z_{c_j}| = 0$, then no schedule $\Phi$ can exist.

This algorithm, based on Propositions 1 and 2, runs in 34716 seconds in Maplesoft (just under 10 hours). Maplesoft generates zero inter-league schedules with $d(\Phi) < 187$ and 14 inter-league schedules with $d(\Phi) = 187$, including the schedule given in Table 8. Since we made the assumption that $p_5$ plays a home set against $c_1$ within the first six time slots, there are actually twice as many distance-optimal schedules by reading each schedule $\Phi$ backwards.

In each of the 28 distance-optimal schedules $\Phi$, we find that $(d(\pi_{p_1}), d(\pi_{p_2}), \ldots, d(\pi_{p_6})) = (0, 4, 0, 0, 1, 1)$ and $(d(\pi_{c_1}), d(\pi_{c_2}), \ldots, d(\pi_{c_6})) = (51, 9, 31, 58, 19, 13)$.

Therefore, we have proven that Table 8 is an optimal inter-league schedule for the NPB, reducing the total travel distance by 8184 kilometres, or 16.0%, compared to the 2010 NPB schedule.

## 5. American Basketball

The National Basketball Association (NBA) is one of the world's most lucrative sports leagues, with over four billion dollars in annual revenue, and an average franchise value of 400 million dollars. There are 15 teams in the Western Conference and 15 teams in the Eastern Conference. Every NBA team plays 82 regular-season games, of which 30 are inter-league (with one home game and one away game against each of the 15 teams from the other conference.) The geographic location of each team is provided in Figure 6.

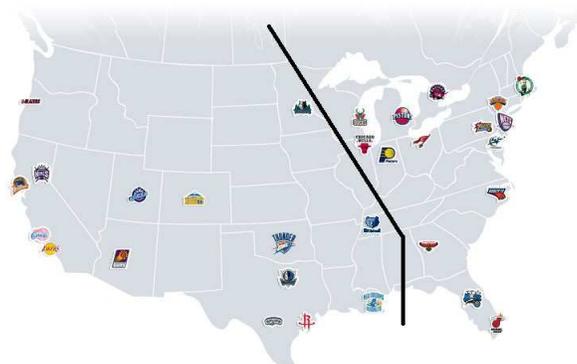

Figure 6: Map of the NBA's 15 Western Conference teams and 15 Eastern Conference teams.

Given that NBA teams play inter-league games, we consider $BTTP$ for this league, where we attempt to find a distance-optimal inter-league tournament. In this theoretical problem, we will assume that all inter-league games take place during a consecutive stretch in the regular season, as is done currently in the Japanese NPB. We will also enforce all the constraints of $BTTP$, including no team having a home stand or road trip lasting more than 3 games. We note that these strict conditions are not part of the NBA scheduling requirement, as evidenced by the San Antonio Spurs playing 6 consecutive home games followed immediately by 8 consecutive road games during the 2009-10 regular season. Furthermore, we will require that our inter-league schedule be *compact*, i.e., having each team play one game in each time slot. Of course, this compactness condition is not part of a typical NBA schedule, as one team might play five games by the time another team has played just two.

We determine the 30 × 30 NBA distance matrix from an online website[1] that lists the flight distance (in statute miles) between each pair of major cities in North America. This matrix is found in Appendix B.

Unlike the 12-team NPB where we could solve $BTTP$, it appears highly unlikely that we can solve this problem for the 30-team NBA. Nonetheless, we can generate an inter-league tournament whose total distance is close to the trivial lower bound of $\sum ILB_t$, by grouping each league's fifteen teams into five triplets so that the travel distance of each team $t$ is extremely close to $ILB_t$, the minimum weight of a $t$-rooted 4-cycle-cover. From this, we construct a uniform tournament, i.e., a feasible solution to $BTTP^*$, where each Western Conference team alternates by playing three away games followed by three home games.

Given the geographic location of the 30 teams, it is easy to show that each team's $ILB_t$ occurs when playing the fifteen away games in five groups of three. We note that this is not always the case; to give a concrete example, consider a variant of the scenario we presented at the end of Section 3. Let points $X, Y, Z$ be equally spaced around a unit circle, so that $\triangle XYZ$ is equilateral. Place $\{e_1, e_2\}$ at $X$, $\{e_3, e_4\}$ at $Y$, and $\{e_5, e_6\}$ at $Z$. Now place $\{w, e_7, e_8, \ldots, e_{15}\}$ at the centre of the circle. Then the best lower bound of $ILB_w = 6$ occurs when $w$ plays two-game road trips against $\{e_1, \ldots, e_6\}$ in pairs rather than three-game road trips like $\{e_1, e_2, e_3\}$ and $\{e_4, e_5, e_6\}$, which has total distance $4 + 2\sqrt{3} > 6$. However, for the NBA distance matrix, each team's $ILB_t$ occurs when that team has five road trips, where in each trip that team plays three opponents located close to each other.

Thus, for each team $w_i$, there exists some permutation $\pi$ for which the lower bound $ILB_{w_i}$ is attained by playing away games against the fifteen Eastern Conference teams in the order $e_{\pi(1)}, e_{\pi(2)}, \ldots, e_{\pi(15)}$. Note that for this permutation, the total distance traveled by $w_i$ is

$$ILB_{w_i} = \sum_{j=1}^{5} \{D_{w_i, e_{\pi(3j-2)}} + D_{e_{\pi(3j-2)}, e_{\pi(3j-1)}} + D_{e_{\pi(3j-1)}, c_{\pi(3j)}} + D_{e_{\pi(3j)}, w_i}\}.$$

The five triplets $\{\{e_{\pi(3j-2)}, e_{\pi(3j-1)}, e_{\pi(3j)}\} : j = 1, 2, \ldots, 5\}$ can be permuted in 5! ways without changing the total distance. Also, within each triplet, we can change the order of the first and third element while retaining the same total. Thus, we can compute $ILB_{w_i}$ from a simple enumeration of $\frac{15!}{5! \cdot 2^5}$ cases, which can be done in minutes using Maplesoft. From this, we calculate $ILB_t$ for each team $t$, giving $LLB_W \geq \sum_{t \in W} ILB_t = 251795$. Similarly, we have $LLB_E \geq \sum_{t \in E} ILB_t = 266137$, and so $TLB \geq LLB_W + LLB_E \geq 517932$.

In nearly every case, the bounds $ILB_{w_i}$ and $ILB_{e_i}$ are attained by selecting the road trips as indicated in Figure 7, corresponding to the minimum-weight *triangle packing* for each league. For example, in this minimum-weight triangle packing, every Eastern Conference team makes just one trip to the northwest, to play Portland, Golden State, and Sacramento in some order. Similarly, every Western Conference team makes just one trip to the southeast, to play Atlanta, Orlando, and Miami in some order. We note the natural connection between minimum-weight triangle packings and minimum-weight 4-cycle-covers, remarking that the former generates an approximation for the latter.

Re-label the fifteen Western Conference teams so that five triplets occur side-by-side (i.e., $w_1$ is Portland, $w_2$ is Golden State, $w_3$ is Sacramento), and similarly re-label the Eastern

---

1. http://www.savvy-discounts.com/discount-travel/JavaAirportCalc.html

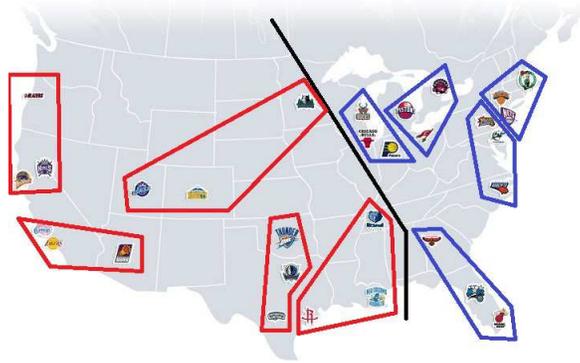

Figure 7: The minimum-weight triangle packing for the 30 NBA teams.

Conference teams. Similar to our construction in Table 5, we build a tournament from the $5 \times 5 = 25$ pairs of inter-league triplets, where each team from one triplet plays the three teams from the other triplet in three consecutive time slots (e.g., $e_1$ plays $\{w_1, w_2, w_3\}$, $e_2$ plays $\{w_2, w_3, w_1\}$ and $e_3$ plays $\{w_3, w_1, w_2\}$.) This construction produces a schedule where the Eastern Conference teams travel 286683 miles and the Western Conference teams travel 258443 miles, for a total of 545126 miles.

We can improve this bound slightly by noting that the *away* teams are not forced to travel according to the triplets given in Figure 7. Specifically, suppose we are considering the road trips for the Eastern Conference. Let the triplets be $\{\{e_{\pi(3j-2)}, e_{\pi(3j-1)}, e_{\pi(3j)}\} : j = 1, 2, \ldots, 5\}$, for some permutation $\pi$. Then each triplet $\{e_{\pi(3j-2)}, e_{\pi(3j-1)}, e_{\pi(3j)}\}$ travels west for three-game road trips against each of $\{w_1, w_2, w_3\}, \{w_4, w_5, w_6\}, \ldots, \{w_{13}, w_{14}, w_{15}\}$. Examining all $\frac{15!}{5! \cdot 2^5}$ non-equivalent possibilities for $\pi$, we show that the best permutation is $\pi = (1, 6, 12, 2, 8, 13, 3, 7, 11, 4, 10, 14, 5, 9, 15)$, so that the teams in $\{e_1, e_6, e_{12}\}$ play their first three games on the road against each of $\{w_1, w_2, w_3\}$, the teams in $\{e_2, e_8, e_{13}\}$ play their first three games on the road against each of $\{w_4, w_5, w_6\}$, and so on. In this optimal schedule, the Eastern Conference teams travel a total of 280294 miles. Similarly, in the best possible case, the Western Conference teams travel a total of 257497 miles.

From this, we produce Table 9, a uniform inter-league tournament with total distance $280294 + 257497 = 537791$ miles, just 3.8% more than the trivial lower bound of $\sum ILB_t$. The labeling of the 30 teams (e.g. PT = Portland Trailblazers, MB = Milwaukee Bucks) is given in Appendix B.

While we are certain that the trivial lower bound of $\sum ILB_t$ cannot be achieved for either the *BTTP* or *BTTP\**, we conjecture that the 3.8% figure can be reduced using more sophisticated techniques. But how close can we get? We leave this as a challenge for the interested reader.

**Problem 1.** *Determine better (best?) bounds for BTTP and BTTP\*, for the $30 \times 30$ NBA distance matrix.*

|    | 1  | 2  | 3  | 4  | 5  | 6  | 7  | 8  | 9  | 10 | 11 | 12 | 13 | 14 | 15 |
|----|----|----|----|----|----|----|----|----|----|----|----|----|----|----|----|
| PT | MB | IP | CU | **MB** | **CB** | **CC** | AH | OM | MH | **CU** | **AH** | **NK** | WW | CB | PS |
| GW | TR | CC | DP | **CC** | **MB** | **CB** | MB | IP | CU | **NK** | **CU** | **AH** | MH | OM | AH |
| SK | NK | NN | BC | **CB** | **CC** | **MB** | TR | CC | DP | **AH** | **NK** | **CU** | MB | IP | CU |
| LC | IP | CU | MB | **NN** | **MH** | **TR** | OM | MH | AH | **CC** | **CB** | **MB** | CB | PS | WW |
| LL | WW | CB | PS | **TR** | **NN** | **MH** | NK | NN | BC | **MB** | **CC** | **CB** | TR | CC | DP |
| PS | BC | NK | NN | **MH** | **TR** | **NN** | CC | DP | TR | **CB** | **MB** | **CC** | IP | CU | MB |
| UJ | OM | MH | AH | **DP** | **PS** | **OM** | PS | WW | CB | **TR** | **NN** | **MH** | BC | NK | NN |
| DN | CC | DP | TR | **OM** | **DP** | **PS** | IP | CU | MB | **MH** | **TR** | **NN** | OM | AH | MH |
| OT | AH | OM | MH | **IP** | **WW** | **BC** | WW | CB | PS | **DP** | **OM** | **PS** | NK | NN | BC |
| SS | CU | MB | IP | **BC** | **IP** | **WW** | MH | AH | OM | **PS** | **DP** | **OM** | PS | WW | CB |
| DM | DP | TR | CC | **WW** | **BC** | **IP** | CU | MB | IP | **OM** | **PS** | **DP** | AH | MH | OM |
| HR | PS | WW | CB | **AH** | **NK** | **CU** | BC | NK | NN | **WW** | **BC** | **IP** | DP | TR | CC |
| MT | CB | PS | WW | **PS** | **OM** | **DP** | NN | BC | NK | **NN** | **MH** | **TR** | CC | DP | TR |
| MG | NN | BC | NK | **CU** | **AH** | **NK** | DP | TR | CC | **IP** | **WW** | **BC** | CU | MB | IP |
| NH | MH | AH | OM | **NK** | **CU** | **AH** | CB | PS | WW | **BC** | **IP** | **WW** | NN | BC | NK |

|    | 16 | 17 | 18 | 19 | 20 | 21 | 22 | 23 | 24 | 25 | 26 | 27 | 28 | 29 | 30 |
|----|----|----|----|----|----|----|----|----|----|----|----|----|----|----|----|
| PT | **BC** | **IP** | **WW** | NK | NN | BC | **DP** | **OM** | **PS** | TR | CC | DP | **TR** | **MH** | **NN** |
| GW | **WW** | **BC** | **IP** | PS | WW | CO | **PS** | **DP** | **OM** | NN | NK | BC | **NN** | **TR** | **MH** |
| SK | **IP** | **WW** | **BC** | AH | OM | MH | **OM** | **PS** | **DP** | WW | CO | PS | **MH** | **NN** | **TR** |
| LC | **NK** | **AH** | **CU** | NN | BC | NK | **IP** | **WW** | **BC** | DP | TR | CC | **DP** | **OM** | **PS** |
| LL | **CU** | **NK** | **AH** | MB | CU | IP | **BC** | **IP** | **WW** | OM | MH | AH | **PS** | **DP** | **OM** |
| PS | **AH** | **CU** | **NK** | MH | AH | OM | **WW** | **BC** | **IP** | PS | WW | CO | **OM** | **PS** | **DP** |
| UJ | **MB** | **CC** | **CO** | CC | DP | TR | **CU** | **NK** | **AH** | CU | IP | MB | **BC** | **WW** | **IP** |
| DN | **CO** | **MB** | **CC** | CO | PS | WW | **AH** | **CU** | **NK** | BC | NN | NK | **IP** | **BC** | **WW** |
| OT | **TR** | **MH** | **NN** | DP | TR | CC | **MB** | **CO** | **CC** | MB | CU | IP | **CU** | **AH** | **NK** |
| SS | **NN** | **TR** | **MH** | BC | NK | NN | **CC** | **MB** | **CO** | CC | DP | TR | **NK** | **CU** | **AH** |
| DM | **MH** | **NN** | **TR** | WW | CO | PS | **CO** | **CC** | **MB** | NK | BC | NN | **AH** | **NK** | **CU** |
| HR | **OM** | **PS** | **DP** | IP | MB | CU | **MH** | **NN** | **TR** | AH | OM | MH | **CO** | **CC** | **MB** |
| MT | **CC** | **CO** | **MB** | CU | IP | MB | **NK** | **AH** | **CU** | MH | AH | OM | **WW** | **IP** | **BC** |
| MG | **DP** | **OM** | **PS** | OM | MH | AH | **TR** | **MH** | **NN** | CO | PS | WW | **MB** | **CO** | **CC** |
| NH | **PS** | **DP** | **OM** | TR | CC | DP | **NN** | **TR** | **MH** | IP | MB | CU | **CC** | **MB** | **CO** |

Table 9: A close-to-optimal solution for the NBA *BTTP\**.

## 6. Conclusion

In this paper, we introduced the Bipartite Traveling Tournament Problem and applied it to two professional sports leagues in Japan and North America, illustrating the richness and complexity of bipartite tournament scheduling.

In Section 4, we introduced two heuristics that enabled us to solve *BTTP* for the $n = 6$ NPB. While Proposition 2 is only applicable for certain 12-team configurations satisfying a specific geometric property, we note that Proposition 1 is a general technique that can be applied to other scheduling problems. Our method of "reduction prior to propagation" breaks a complex problem into a large number of scenarios, and sets up each scenario as a global constraint to reduce the search space. We are confident that Proposition 1 can be applied to more complicated problems in sports scheduling.

In Section 5, we determined an algorithm that produced an approximate solution to *BTTP* for the $n = 15$ NBA. By finding minimum-weight rooted 4-cycle-covers, we determined a trivial lower bound to *BTTP*, from which our method of creating a uniform schedule based on the minimum-weight triangle packing generated a close-to-optimal feasible solution. For the NBA inter-league problem, this process produced an optimality gap of just 3.8%. We are hopeful that these ideas can be abstracted and refined further, leading to more powerful tools to tackle even harder problem instances.

Perhaps there are other sports leagues for which *BTTP* is applicable, such as in professional hockey and football. We can also expand our analysis to model *tripartite* and *multipartite* tournament scheduling, where a league is divided into three or more conferences. A specific example of this is the newly-created Super 15 Rugby League, consisting of five teams from South Africa, Australia, and New Zealand. In addition to intra-country games, each team plays four games (two home and two away) against teams from each of the other two countries. It would be interesting to see whether we can determine the distance-optimal tripartite tournament schedule using the methods developed in this paper.

We conclude by motivating several interesting questions, including those dealing with geometric probability and extremal combinatorics, and leave them as open problems for the interested reader.

Our solution to the non-uniform *BTTP* required 10 hours of computations. Furthermore, we were only able to solve *BTTP* by applying Proposition 2, whose requirements would not hold for a randomly-selected $12 \times 12$ distance matrix. As a result, we require a more sophisticated technique that improves upon our two heuristics, perhaps using methods in constraint programming and integer programming, such as a hybrid CP/IP. We wonder if there exists a general algorithm that would solve *BTTP* given any distance matrix, for "small" values of $n$ such as $n = 6$, $n = 7$, and $n = 8$. We pose this as an open problem.

**Problem 2.** *Develop a computational procedure (or algorithm) that can routinely solve BTTP and BTTP\* instances with $n \geq 6$.*

At the end of Section 3, we presented a simple example (see Figure 4) to illustrate the difference between *BTTP* and *BTTP\** for the case $n = 3$. We located the six points to form two sets of Pythagorean triangles, and showed that the solutions to the two problems had total distance $15a + 3b + 5c$ and $16a + 4b + 4c$, respectively. If $(a, b, c) = (3, 4, 5)$, then the tournament lower bounds are 82 and 84, respectively. In other words, by relaxing the uniformity requirement, we can reduce the optimal travel distance from 84 to 82, an improvement of 2.38%. Using elementary calculus, we can show that for this particular choice of six points, the percentage reduction function is at most 2.39%, with equality iff $\frac{b}{a} = \frac{5+3\sqrt{5}}{8}$. However, if we selected a different set of six points, could we achieve a better percentage reduction? This motivates the following question:

**Problem 3.** *Consider six points $X = \{x_1, x_2, x_3\}$ and $Y = \{y_1, y_2, y_3\}$ in the Cartesian plane. Let $D^*$ and $D$ be the tournament lower bounds of BTTP\* and BTTP, respectively. Determine the smallest constant $c$ for which $D^* \leq c \cdot D$ for all possible selections of the six points in $X \cup Y$.*

One may conjecture that in order to minimize the tournament lower bound, we must minimize the total number of trips taken by the $2n$ teams. But as we saw in Table 6, this conjecture is false for $n = 3$, as we located six points for which the solution to *BTTP\** requires 24 trips, while the solution to *BTTP* requires 25 trips. However, there are numerous examples (e.g. our scenario in Table 7) where the $2n$ points can be located so that the solution to *BTTP\** matches that of *BTTP*. This motivates the following question: given a random selection of $2n$ points, what is the probability that the solutions to *BTTP\** and *BTTP* are identical?

To illustrate, consider the case $n = 3$. We can quickly show that there exist $60 \times 2^9 = 30720$ feasible inter-league tournaments, of which $60 \times 2^3 = 480$ are uniform. We run a simulation on Maplesoft, where in each scenario, we randomly select six points $(x, y)$ in the Cartesian plane, calculate the $15 \times 1$ column vector of pairwise distances, and apply it to the set of feasible inter-league tournaments to determine the distance-optimal schedule. We run the simulation 100000 times, where in each scenario, we note the number of trips taken in the optimal solution. The results appear in Table 10.

| Trips | 24 | 25 | 26 | 27 | 28 | $\geq 29$ |
|---|---|---|---|---|---|---|
| Scenarios | 55800 | 33077 | 10967 | 43 | 0 | 0 |

Table 10: Results of simulation: number of trips in the distance-optimal tournament.

We note that the sum total is not 100000, as there were 113 scenarios that ended in a tie (e.g. there were two tournaments, one with 24 trips and another with 26 trips, both with equal total distance after rounding to two decimal places.)

As expected, in the majority of scenarios, the six points $X \cup Y$ had the property that the distance-optimal bipartite tournament involved 24 trips, where each team played three consecutive road games. Without much difficulty, one can show (Hoshino & Kawarabayashi, 2011a) that this forces the home game slots to be *uniform*, i.e., all the teams in each league must play their home games at the same time, and so each team in $X$ plays three consecutive home games followed by three consecutive road games, or vice-versa. Therefore, in 55.8% of our randomly-selected scenarios, the solution to *BTTP\** is identical to the solution to *BTTP*.

Our simulation motivates an interesting question in geometric probability. Given that the $2n$ points of $X \cup Y$ are chosen at random, what is the probability that the tournament lower bound is achieved with a schedule consisting of $t$ trips? We formally define the question below and present it as an open problem for the reader.

**Problem 4.** *Let the $2n$ points $X = \{x_1, x_2, \ldots, x_n\}$ and $Y = \{y_1, y_2, \ldots, y_n\}$ be randomly selected in the Cartesian plane. Let $t$ be the number of trips taken in a distance-optimal solution of BTTP, with the teams located at $X$ and $Y$. Determine the value $P_n(t)$ for each $t$, where $P_n(t)$ represents the probability that the distance-optimal tournament involves the $2n$ teams taking exactly $t$ trips.*

For the case $n = 3$, it appears that $P_n(t) = 0$ for all $t \leq 23$ and $t \geq 28$. While it is trivial to show that we must have at least 24 trips, we do not have a formal proof that there cannot exist a selection of six points $X \cup Y$ in the plane for which the solution to *BTTP* has more than 27 trips. If we could prove that for each $n$, the number of total trips in a distance-optimal solution is bounded above by some function $f(n)$, then this would enable us to solve *BTTP* without having to enumerate all feasible schedules, i.e., a small fraction would suffice. Such a result would certainly aid in solving *BTTP* for larger $n$, where a full enumeration of all feasible tournament schedules is too computationally laborious. This motivates our final problem.

**Problem 5.** *Consider a $2n$-team bipartite tournament, with the teams located at $X = \{x_1, x_2, \ldots, x_n\}$ and $Y = \{y_1, y_2, \ldots, y_n\}$. For each $n$, determine the smallest integer $f(n)$* *for which the solution to BTTP involves the teams taking at most $f(n)$ trips, regardless of where the $2n$ teams are located.*

## Acknowledgments

This research has been partially supported by the Japan Society for the Promotion of Science (Grant-in-Aid for Scientific Research), the C & C Foundation, the Kayamori Foundation, and the Inoue Research Award for Young Scientists.

## Appendix A.

We provide the proof of our three lemmas (from Section 3), beginning with Lemma 1.

*Proof.* First, we prove $(i) \to (ii)$.

If $S$ is satisfiable, then there exists a function $\phi$ that is a valid truth assignment, i.e., a function for which $\phi(u_i) \in \{\text{TRUE}, \text{FALSE}\}$ for each $1 \leq i \leq l$ that ensures that each clause $C_j$ evaluates to TRUE for all $1 \leq j \leq 2k$. From $\phi$, we build a $p$-rooted 4-cycle-cover of $K_S$ with exactly $3k$ $(p,a,u,b,p)$-cycles, $2k$ $(p,u,c,d,p)$-cycles, and $k$ $(p,u,e,f,p)$-cycles.

We first identify the $3k$ $(p,a,u,b,p)$-cycles. For each $1 \leq i \leq l$, if $\phi(u_i)$ is FALSE, then select all 4-cycles of the form $p$-$a_{i,r}$-$u_{i,r}$-$b_{i,r}$-$p$, for each $r = 1, 2, \ldots, r(i)$. And if $\phi(u_i)$ is TRUE, then select all 4-cycles of the form $p$-$a_{i,r+1}$-$\overline{u}_{i,r}$-$b_{i,r}$-$p$, for each $r$ (where $a_{i,r(i)+1} = a_{i,1}$). Repeating this construction for each $i$, we produce $3k$ $(p,a,u,b,p)$-cycles, covering the $6k$ vertices of $A \cup B$, as well as $3k$ vertices of $U$.

Now consider any clause $C_j$. Since $\phi$ is a valid truth assignment, at least one of the three literals in $C_j$ evaluates to TRUE. In other words, there must exist some index $i$ for which $u_i \in C_j$ and $\phi(u_i)$ is TRUE, or $\overline{u}_i \in C_j$ and $\phi(u_i)$ is FALSE.

In the former case, where $u_i \in C_j$ and $\phi(u_i)$ is TRUE, there exists some index $r$ for which $u_{i,r}$-$c_j$ is an edge of the gadget graph $G_S$. Then $p$-$u_{i,r}$-$c_j$-$d_j$-$p$ is a $(p,u,c,d,p)$-cycle. Note that $u_{i,r}$ has not been previously selected in a $(p,a,u,b,p)$-cycle since $\phi(u_i)$ is TRUE (and so only the vertices $\overline{u}_{i,1}, \overline{u}_{i,2}, \ldots, \overline{u}_{i,r(i)}$ were covered earlier.)

In the latter case, where $\overline{u}_i \in C_j$ and $\phi(u_i)$ is FALSE, there exists some index $r$ for which $\overline{u}_{i,r}$-$c_j$ is an edge of the gadget graph $G_S$. Then $p$-$\overline{u}_{i,r}$-$c_j$-$d_j$-$p$ is a $(p,u,c,d,p)$-cycle. Note that $\overline{u}_{i,r}$ has not been previously selected in a $(p,a,u,b,p)$-cycle since $\phi(u_i)$ is FALSE (and so only the vertices $u_{i,1}, u_{i,2}, \ldots, u_{i,r(i)}$ were covered earlier.)

Repeating this construction for each $j$, we produce $2k$ $(p,u,c,d,p)$-cycles, covering the $4k$ vertices of $C \cup D$. Note that no $u \in U$ can be chosen twice since each vertex in $U$ is adjacent to only one vertex in $C$. Thus, these $2k$ cycles cover a set of $6k$ vertices in $X_S$, completely disjoint from the $9k$ vertices covered by the previously-constructed $3k$ $(p,a,u,b,p)$-cycles. As a result, we are left with $3k$ vertices in $X_S$ still to be covered, specifically $k$ vertices in each of $U$, $E$, and $F$. These vertices can be trivially partitioned into $k$ $(p,u,e,f,p)$-cycles by just ensuring that $e_j$ and $f_j$ belong to the same cycle for each $1 \leq j \leq k$. When this process is complete, our $p$-rooted 4-cycle-cover of $K_S$ will contain exactly $3k$ $(p,a,u,b,p)$-cycles, $2k$ $(p,u,c,d,p)$-cycles, and $k$ $(p,u,e,f,p)$-cycles.

Having established the first direction, we now prove $(ii) \to (i)$.

Consider a $p$-rooted 4-cycle-cover of $K_S$ containing exactly $3k$ $(p, a, u, b, p)$-cycles, $2k$ $(p, u, c, d, p)$-cycles, and $k$ $(p, u, e, f, p)$-cycles. We prove there exists a function $\phi$ that is a satisfying truth assignment for $S$, where $\phi(u_i) \in \{\text{TRUE}, \text{FALSE}\}$ for each $1 \leq i \leq l$.

Define an *a-b path* to be any path on three vertices whose endpoints are $a_{i,j}$ and $b_{i,k}$, for some indices $i, j, k$. Consider the problem of maximizing the number of vertex-disjoint *a-b* paths in the $i^\text{th}$ gadget. One can quickly see that a maximum packing of *a-b* paths occurs iff the $r(i)$ paths are chosen in one of the following "trivial" ways:

(a) Taking all paths of the form $a_{i,r}, u_{i,r}, b_{i,r}$ for each $r = 1, 2, \ldots, r(i)$.

(b) Taking all paths of the form $a_{i,r+1}, \overline{u}_{i,r}, b_{i,r}$ for each $r = 1, 2, \ldots, r(i)$.   (Note: $a_{i,r(i)+1} = a_{i,1}$.)

In order for us to cover all of the vertices in $A \cup B$, in each gadget we must select our *a-b* paths either vertically (a) or diagonally (b). Thus, in our $p$-rooted 4-cycle-cover containing $3k$ $(p, a, u, b, p)$-cycles, one of the following scenarios must hold true in the $i^\text{th}$ gadget:

(1) For each $r = 1, 2, \ldots, r(i)$, vertex $u_{i,r}$ appears in some $(p, a, u, b, p)$-cycle, while no vertex $\overline{u}_{i,r}$ appears in any $(p, a, u, b, p)$-cycle.

(2) For each $r = 1, 2, \ldots, r(i)$, vertex $\overline{u}_{i,r}$ appears in some $(p, a, u, b, p)$-cycle, while no vertex $u_{i,r}$ appears in any $(p, a, u, b, p)$-cycle.

In our given $p$-rooted 4-cycle-cover of $K_S$, for each $i$ define $\phi(u_i) = \text{FALSE}$ in scenario (1) and define $\phi(u_i) = \text{TRUE}$ in scenario (2). We claim that this is our desired function $\phi$.

To prove this, consider the $2k$ $(p, u, c, d, p)$-cycles in our 4-cycle-cover. For each $1 \leq j \leq 2k$, the $(p, u, c, d, p)$-cycle containing $c_j$ also contains some other vertex in $U$. This vertex is either $u_{i,r}$ or $\overline{u}_{i,r}$, for some indices $i$ and $r$.

In the former case, $u_{i,r}$ and $c_j$ appear in the same $(p, u, c, d, p)$-cycle, implying that $u_{i,r}$-$c_j$ is an edge of the gadget graph $G_S$, and that $u_i$ is a literal in clause $C_j$. Since $u_{i,r}$ appears in this $(p, u, c, d, p)$-cycle and therefore not in any $(p, a, u, b, p)$-cycle, this implies scenario (2) above. Since $\phi(u_i) = \text{TRUE}$ and $u_i \in C_j$, clause $C_j$ evaluates to TRUE.

In the latter case, $\overline{u}_{i,r}$ and $c_j$ appear in the same $(p, u, c, d, p)$-cycle, implying that $\overline{u}_{i,r}$-$c_j$ is an edge of the gadget graph $G_S$, and that $\overline{u}_i$ is a literal in clause $C_j$. Since $\overline{u}_{i,r}$ appears in this $(p, u, c, d, p)$-cycle and therefore not in any $(p, a, u, b, p)$-cycle, this implies scenario (1) above. Since $\phi(u_i) = \text{FALSE}$ and $\overline{u}_i \in C_j$, clause $C_j$ evaluates to TRUE.

Since $C_j$ evaluates to TRUE for all $1 \leq j \leq 2k$, this implies that $\phi$ is a valid truth assignment. We conclude that $S = C_1 \wedge C_2 \wedge \ldots \wedge C_{2k}$ is satisfiable. $\square$

We now prove Lemma 2.

*Proof.* First, we prove $(i) \rightarrow (ii)$.

In a $(p, a, u, b, p)$-cycle, the edges $au$ and $ub$ appear in the gadget graph $G_S$. Therefore, the edge weights of $au$ and $ub$ are both $z^2$. From Table 3, we see that a $(p, a, u, b, p)$-cycle has edge weight $z^2 + z^2 + z^2 + z^2 = 4z^2$. Similarly, a $(p, u, c, d, p)$-cycle has edge weight $(z^2+z)+z^2+z^2+z^2 = 4z^2+z$, and a $(p, u, e, f, p)$-cycle has edge weight $(z^2+z)+z^2+z^2+z^2 = 4z^2 + z$.

So if a $p$-rooted 4-cycle-cover of $K_S$ has exactly $3k$ $(p, a, u, b, p)$-cycles, $2k$ $(p, u, c, d, p)$-cycles, and $k$ $(p, u, e, f, p)$-cycles, then its total edge weight is exactly $3k(4z^2) + 2k(4z^2 + z) + k(4z^2 + z) = k(24z^2 + 3z)$.

Having established the first direction, we now prove $(ii) \to (i)$.

Let $R$ be a $p$-rooted 4-cycle-cover of $K_S$ which is the union of $r$ cycles, with total edge weight $k(24z^2 + 3z)$. Since each of the 18k vertices of $X_S$ is covered by exactly one cycle of $R$, the number of edges in $R$ is $|X_S| + r = 18k + r$. Since no cycle has length greater than 4, we have $r \geq \frac{18k}{3} = 6k$. Now suppose $r \geq 6k + 1$. Then there are at least $24k + 1$ edges in $R$, all of which have weight at least $z^2$ given the construction of our complete graph $K_S$. Hence, the total edge weight of $R$ is at least $(24k+1)z^2 = 24kz^2 + z^2 = 24kz^2 + z(20k+1) > 24kz^2 + 3zk = k(24z^2 + 3z)$, a contradiction.

It follows that $r = 6k$, and that $R$ must be the union of $6k$ cycles of length 4. Recall that the weight of each edge appears in the set $\{z^2, z^2 + z, z^2 + 2z, 2z^2 - 1\}$. Suppose that one of these $24k$ edges has weight $2z^2 - 1$. Then the total edge weight of $R$ is at least $(24k-1)z^2 + (2z^2 - 1) = 24kz^2 + z^2 - 1 = 24kz^2 + z(20k+1) - 1 > 24kz^2 + 3zk = k(24z^2 + 3z)$, a contradiction. Hence, all edges of $R$ must have weight $z^2$, $z^2 + z$, or $z^2 + 2z$.

From Table 3, we see that no edges $p$-$c$ and $p$-$e$ can appear in our 4-cycle-cover $R$, since all edges from $p$ to $C \cup E$ have weight $2z^2 - 1$. It follows that there must exist $2k$ 4-cycles of the form $p$-?-$c_i$-?-$p$ and $k$ 4-cycles of the form $p$-?-$e_i$-?-$p$, with each of these $2k + k = 3k$ 4-cycles containing a unique element from $C \cup E$. Each blank space (denoted by a question mark) can only be filled with a vertex from $D$, $F$, or $U$, as the weights of edges $ca$, $cb$, $ea$, $eb$ are all $2z^2 - 1$ for all $a \in A$, $b \in B$, $c \in C$, and $e \in E$.

Since edge $p$-$u$ has weight $z^2 + z$, if some vertex $u \in U$ is chosen to appear in one of these $3k$ 4-cycles, then this adds edge weight $z^2 + z$, producing a 4-cycle of weight at least $4z^2 + z$. But if no vertices $u \in U$ are chosen to replace these blank spaces, then the cycles must be of the form $p$-$d_j$-$c_i$-$d_k$-$p$ or $p$-$f_j$-$e_i$-$f_k$-$p$, both of which lead to the addition of at least one edge of weight $2z^2 - 1$ (since we cannot simultaneously have $i = j$, $i = k$, and $j \neq k$). It follows that these $2k + k = 3k$ 4-cycles containing the vertices of $C \cup E$ must each have weight at least $4z^2 + z$, thus contributing at least $k(12z^2 + 3z)$ to the total distance of the $p$-rooted 4-cycle-cover $R$.

Since the given 4-cycle-cover $R$ has weight $k(24z^2 + 3z)$, this implies that the rest of the $3k$ 4-cycles must each have weight exactly $4z^2$, and that in each of the $2k$ cycles of the form $p$-?-$c_i$-?-$p$ and $k$ cycles of the form $p$-?-$e_i$-?-$p$, the total edge weight must be *exactly* $4z^2 + z$ to ensure that the total edge weight of $R$ does not exceed $k(24z^2 + 3z)$. This implies that in these two scenarios, we cannot replace the two blank spaces with two distinct vertices from $U$, as that would create a cycle of weight $4z^2 + 2z$. It follows that $R$ *must* have $2k$ $(p, u, c, d, p)$-cycles and $k$ $(p, u, e, f, p)$-cycles.

We are now left with $3k$ vertices from each of $A$, $B$, and $U$ to form our remaining $3k$ 4-cycles. In order for the total edge weight of $R$ to not exceed $k(24z^2 + 3z) = 3k(4z^2 + z) + 12kz^2$, each of the remaining $12k$ edges must have weight $z^2$. Since edge $p$-$u$ has weight $z^2 + z$ for all $u \in U$, the $3k$ remaining vertices in $U$ must each appear in a unique 4-cycle, none adjacent to the root vertex $p$. It follows that the remaining $3k$ 4-cycles of $R$ must all be $(p, a, u, b, p)$-cycles. □

We now prove Lemma 3.

*Proof.* From the proof of Lemma 2, we see that $ILB_p = k(24z^2 + 3z)$, so this handles the case $y = p$. We now consider the case $y = q$.

Let $R$ be a $q$-rooted 4-cycle-cover of $K_S$ which is the union of $r$ cycles. Suppose on the contrary that there exists an $R$ for which its total edge weight is less than $k(24z^2 + 20z)$. We will derive a contradiction.

Since each of the 18k vertices of $X_S$ is covered by exactly one cycle of $R$, the number of edges in $R$ is $|X_S| + r = 18k + r$. As in the previous proof, $r \geq 6k$. If $r \geq 6k+1$, then the total edge weight of $R$ is at least $(24k+1)z^2 = 24kz^2 + z^2 = 24kz^2 + z(20k+1) > k(24z^2 + 20z)$, a contradiction.

Hence, $r = 6k$, and so $R$ must be the union of $6k$ cycles of length 4. Now suppose that one of these $24k$ edges has weight $2z^2 - 1$. Then the total edge weight of $R$ is at least $(24k-1)z^2 + (2z^2 - 1) = 24kz^2 + z^2 - 1 = 24kz^2 + z(20k+1) - 1 > k(24z^2 + 20z)$, another contradiction.

Therefore, no edges $q$-$b$ and $q$-$d$ can appear in our 4-cycle-cover $R$, since all edges from $p$ to $B \cup D$ have weight $2z^2 - 1$. It follows that there must exist $3k$ 4-cycles of the form $q$-?-$b_i$-?-$q$ and $2k$ 4-cycles of the form $q$-?-$d_i$-?-$q$. No blank space (denoted by a question mark) can be filled with a vertex from $E$ or $F$ since the weights of edges $be$, $bf$, $de$, $df$ are $2z^2 - 1$ for all $b \in B$, $d \in D$, $e \in E$, and $f \in F$.

It follows that the $k$ remaining 4-cycles must include all of the $k + k = 2k$ vertices in $E \cup F$. If any of these 4-cycles contains three elements of $E \cup F$ (e.g. the cycle $q$-$e_i$-$f_j$-$e_k$-$q$ or the cycle $q$-$e_i$-$e_j$-$f_k$-$q$), then that creates at least one edge with weight $2z^2 - 1$, a contradiction. Thus, there must be exactly two vertices from $E \cup F$ in each of these 4-cycles. Moreover, since the weights of edges $ae$, $af$, $ce$, $cf$ are all $2z^2 - 1$ for all $a \in A, c \in C$, it follows that the final vertex of these remaining $k$ 4-cycles must be an element of $U$, thus producing a 4-cycle such as $q$-$u_{i,r}$-$e_j$-$f_k$-$q$ or $q$-$f_i$-$u_{j,r}$-$f_k$. From Table 3, we see that every valid cycle has edge weight $\geq 4z^2 + 2z$.

Hence, we must have $k$ 4-cycles in the cycle cover $R$, containing all $2k$ vertices in $E \cup F$ and $k$ vertices in $U$, contributing total weight $\geq k(4z^2 + 2z)$. Of the $3k$ 4-cycles of the form $q$-?-$b_i$-?-$q$, no vertex in $C$ can appear, as otherwise there would be an edge with weight $2z^2 - 1$. Similarly, of the $2k$ 4-cycles of the form $q$-?-$d_i$-?-$q$, no vertex in $A$ can appear.

Thus, in each of the $3k$ 4-cycles containing $b_i$, the other two vertices must be selected from $A \cup U$. From Table 3, we see that every such 4-cycle has edge weight $\geq 4z^2 + 4z$. And in each of the $2k$ 4-cycles containing $d_i$, the other two vertices must be selected from $C \cup U$. Also from Table 3, we see that every such 4-cycle has edge weight $\geq 4z^2 + 3z$.

Therefore, any $q$-rooted 4-cycle-cover of $K_S$ has total edge weight $\geq k(4z^2 + 2z) + 3k(4z^2 + 4z) + 2k(4z^2 + 3z) = k(24z^2 + 20z)$, establishing our desired contradiction. We conclude that $ILB_q = k(24z^2 + 20z)$.

The proof for the $r$-rooted 4-cycle-cover is identical. We just apply the mapping $\{a, b, c, d, e, f, u\} \to \{b, a, e, f, c, d, u\}$ to the vertices in the preceding paragraphs to reach the same conclusion. In this case, we have $ILB_r = k(4z^2+3z)+3k(4z^2+4z)+2k(4z^2+2z) = k(24z^2 + 19z)$. □

## Appendix B.

We now provide the $12 \times 12$ distance matrix for the NPB league (from Section 4), and the $30 \times 30$ distance matrix for the NBA (from Section 5).

As mentioned in Section 4, the Pacific League teams are $p_1$ (Fukuoka), $p_2$ (Orix), $p_3$ (Saitama), $p_4$ (Chiba), $p_5$ (Tohoku), $p_6$ (Hokkaido), and the Central League teams are $c_1$ (Hiroshima), $c_2$ (Hanshin), $c_3$ (Chunichi), $c_4$ (Yokohama), $c_5$ (Yomiuri), and $c_6$ (Yakult). In Table 11, we only provide $D_{c_i,c_j}$ and $D_{p_i,p_j}$ for $i < j$ since the case $i > j$ is equivalent by symmetry.

| Team | $c_1$ | $c_2$ | $c_3$ | $c_4$ | $c_5$ | $c_6$ | $p_1$ | $p_2$ | $p_3$ | $p_4$ | $p_5$ | $p_6$ |
|---|---|---|---|---|---|---|---|---|---|---|---|---|
| $c_1$ | 0 | 323 | 488 | 808 | 827 | 829 | 258 | 341 | 870 | 857 | 895 | 1288 |
| $c_2$ |   | 0 | 195 | 515 | 534 | 536 | 577 | 27 | 577 | 564 | 654 | 1099 |
| $c_3$ |   |   | 0 | 334 | 353 | 355 | 742 | 213 | 396 | 383 | 511 | 984 |
| $c_4$ |   |   |   | 0 | 37 | 35 | 916 | 533 | 63 | 58 | 364 | 886 |
| $c_5$ |   |   |   |   | 0 | 7 | 926 | 552 | 51 | 37 | 331 | 896 |
| $c_6$ |   |   |   |   |   | 0 | 923 | 554 | 48 | 39 | 333 | 893 |
| $p_1$ |   |   |   |   |   |   | 0 | 595 | 958 | 934 | 1100 | 1466 |
| $p_2$ |   |   |   |   |   |   |   | 0 | 595 | 582 | 670 | 1115 |
| $p_3$ |   |   |   |   |   |   |   |   | 0 | 86 | 374 | 928 |
| $p_4$ |   |   |   |   |   |   |   |   |   | 0 | 361 | 904 |
| $p_5$ |   |   |   |   |   |   |   |   |   |   | 0 | 580 |
| $p_6$ |   |   |   |   |   |   |   |   |   |   |   | 0 |

Table 11: Distance Matrix for the Japanese NPB League.

To calculate each entry of this distance matrix, we determined how the teams travel from one stadium to another, taking into account the actual mode(s) of transportation. For example, the distance $D_{c_2,c_5} = 534$ was found by adding the travel distance for each component of the trip from Hanshin's home stadium to Yomiuri's home stadium, namely the 15 km bus ride from Koshien Stadium to Shin-Osaka Station, the 515 km bullet-train ride to Tokyo Station, followed by the 4 km bus ride to the Tokyo Dome. This is a more rigorous approach that simply calculating the flight distance between the airports in Osaka and Tokyo. Noting when teams travel by airplane, bullet train, and bus, we repeat the analysis for each of the $\binom{12}{2} = 66$ pairs of cities to produce the matrix in Table 11.

Finally, we provide the $30 \times 30$ distance matrix for the NBA, as well the labeling of the 30 teams in Table 9. There are fifteen teams in the Western Conference, namely the Portland Trailblazers (PT), Golden State Warriors (GW), Sacramento Kings (SK), Los Angeles Clippers (LC), Los Angeles Lakers (LL), Phoenix Suns (PS), Utah Jazz (UJ), Denver Nuggets (DN), Oklahoma Thunder (OT), San Antonio Spurs (SS), Dallas Mavericks (DM), Houston Rockets (HR), Minnesota Timberwolves (MT), Memphis Grizzlies (MG), and New Orleans Hornets (NH).

There are fifteen teams in the Eastern Conference, namely the Milwaukee Bucks (MB), Chicago Bulls (CU), Indiana Pacers (IP), Detroit Pistons (DP), Toronto Raptors (TR), Cleveland Cavaliers (CC), Boston Celtics (BC), New York Knicks (NK), New Jersey Nets (NN), Philadelphia Sixers (PS), Washington Wizards (WW), Charlotte Bobcats (CB), Atlanta Hawks (AH), Orlando Magic (OM), and Miami Heat (MH). Note that two teams (Chicago Bulls and Charlotte Bobcats) have the same initials, and thus we have represented the former as CU and the latter as CB to avoid ambiguity.

| Team | PT | GW | SK | LC | LL | PS | UJ | DN | OT | SS | DM | HR | MT | MG | NH |
|---|---|---|---|---|---|---|---|---|---|---|---|---|---|---|---|
| PT | 0 | 536 | 473 | 824 | 824 | 997 | 620 | 969 | 1462 | 1691 | 1602 | 1799 | 1403 | 1826 | 2020 |
| GW |   | 0 | 75 | 333 | 333 | 636 | 580 | 931 | 1353 | 1455 | 1445 | 1605 | 1555 | 1770 | 1875 |
| SK |   |   | 0 | 368 | 368 | 637 | 524 | 884 | 1320 | 1441 | 1420 | 1586 | 1494 | 1733 | 1850 |
| LC |   |   |   | 0 | 0 | 365 | 582 | 837 | 1169 | 1192 | 1227 | 1358 | 1514 | 1594 | 1645 |
| LL |   |   |   |   | 0 | 365 | 582 | 837 | 1169 | 1192 | 1227 | 1358 | 1514 | 1594 | 1645 |
| PS |   |   |   |   |   | 0 | 501 | 582 | 820 | 831 | 866 | 994 | 1258 | 1244 | 1281 |
| UJ |   |   |   |   |   |   | 0 | 375 | 852 | 1072 | 985 | 1178 | 976 | 1242 | 1408 |
| DN |   |   |   |   |   |   |   | 0 | 493 | 785 | 645 | 853 | 683 | 867 | 1052 |
| OT |   |   |   |   |   |   |   |   | 0 | 402 | 178 | 391 | 686 | 425 | 559 |
| SS |   |   |   |   |   |   |   |   |   | 0 | 244 | 187 | 1084 | 617 | 486 |
| DM |   |   |   |   |   |   |   |   |   |   | 0 | 215 | 843 | 417 | 430 |
| HR |   |   |   |   |   |   |   |   |   |   |   | 0 | 1023 | 462 | 300 |
| MT |   |   |   |   |   |   |   |   |   |   |   |   | 0 | 692 | 1027 |
| MG |   |   |   |   |   |   |   |   |   |   |   |   |   | 0 | 345 |
| NH |   |   |   |   |   |   |   |   |   |   |   |   |   |   | 0 |

Table 12: Distance Matrix for the NBA Western Conference (intra-league).

| Team | MB | CU | IP | DP | TR | CC | BC | NK | NN | PS | WW | CB | AH | OM | MH |
|---|---|---|---|---|---|---|---|---|---|---|---|---|---|---|---|
| MB | 0 | 66 | 235 | 248 | 414 | 323 | 847 | 734 | 714 | 680 | 602 | 642 | 661 | 1047 | 1244 |
| CU |   | 0 | 175 | 249 | 430 | 310 | 853 | 728 | 708 | 667 | 580 | 591 | 599 | 985 | 1183 |
| IP |   |   | 0 | 249 | 433 | 257 | 805 | 655 | 634 | 578 | 468 | 422 | 427 | 811 | 1009 |
| DP |   |   |   | 0 | 190 | 90 | 605 | 486 | 466 | 434 | 372 | 502 | 602 | 950 | 1143 |
| TR |   |   |   |   | 0 | 191 | 439 | 361 | 343 | 342 | 341 | 582 | 731 | 1036 | 1220 |
| CC |   |   |   |   |   | 0 | 553 | 418 | 398 | 357 | 284 | 425 | 548 | 877 | 1068 |
| BC |   |   |   |   |   |   | 0 | 184 | 198 | 276 | 407 | 718 | 933 | 1101 | 1243 |
| NK |   |   |   |   |   |   |   | 0 | 20 | 93 | 225 | 534 | 749 | 926 | 1077 |
| NN |   |   |   |   |   |   |   |   | 0 | 80 | 209 | 522 | 735 | 919 | 1074 |
| PS |   |   |   |   |   |   |   |   |   | 0 | 133 | 442 | 657 | 844 | 1002 |
| WW |   |   |   |   |   |   |   |   |   |   | 0 | 317 | 526 | 742 | 911 |
| CB |   |   |   |   |   |   |   |   |   |   |   | 0 | 224 | 456 | 643 |
| AH |   |   |   |   |   |   |   |   |   |   |   |   | 0 | 392 | 589 |
| OM |   |   |   |   |   |   |   |   |   |   |   |   |   | 0 | 198 |
| MH |   |   |   |   |   |   |   |   |   |   |   |   |   |   | 0 |

Table 13: Distance Matrix for the NBA Eastern Conference (intra-league).

| Team | PT | GW | SK | LC | LL | PS | UJ | DN | OT | SS | DM | HR | MT | MG | NH |
|---|---|---|---|---|---|---|---|---|---|---|---|---|---|---|---|
| MB | 1690 | 1806 | 1750 | 1730 | 1730 | 1439 | 1227 | 894 | 726 | 1082 | 840 | 973 | 292 | 550 | 893 |
| CU | 1711 | 1807 | 1753 | 1718 | 1718 | 1418 | 1230 | 886 | 683 | 1028 | 787 | 915 | 330 | 485 | 827 |
| IP | 1848 | 1903 | 1855 | 1786 | 1786 | 1466 | 1333 | 973 | 678 | 973 | 745 | 834 | 495 | 376 | 699 |
| DP | 1934 | 2052 | 1998 | 1967 | 1967 | 1665 | 1474 | 1135 | 908 | 1220 | 990 | 1083 | 531 | 624 | 935 |
| TR | 2064 | 2214 | 2157 | 2143 | 2143 | 1848 | 1634 | 1307 | 1098 | 1406 | 1177 | 1264 | 667 | 801 | 1097 |
| CC | 2014 | 2117 | 2064 | 2022 | 2022 | 1711 | 1540 | 1194 | 934 | 1224 | 1001 | 1077 | 612 | 614 | 906 |
| BC | 2497 | 2651 | 2594 | 2572 | 2572 | 2265 | 2072 | 1739 | 1482 | 1739 | 1532 | 1575 | 1106 | 1123 | 1349 |
| NK | 2415 | 2535 | 2482 | 2437 | 2437 | 2120 | 1958 | 1612 | 1324 | 1564 | 1362 | 1397 | 1012 | 950 | 1166 |
| NN | 2395 | 2515 | 2461 | 2417 | 2417 | 2100 | 1937 | 1592 | 1305 | 1547 | 1344 | 1380 | 992 | 932 | 1151 |
| PS | 2368 | 2472 | 2419 | 2365 | 2365 | 2403 | 1895 | 1544 | 1242 | 1474 | 1275 | 1306 | 965 | 861 | 1074 |
| WW | 2291 | 2372 | 2320 | 2252 | 2252 | 1926 | 1799 | 1441 | 1118 | 1342 | 1147 | 1173 | 894 | 731 | 942 |
| CB | 2247 | 2251 | 2209 | 2092 | 2092 | 1747 | 1700 | 1327 | 926 | 1080 | 912 | 899 | 917 | 503 | 642 |
| AH | 2140 | 2097 | 2060 | 1917 | 1917 | 1562 | 1565 | 1190 | 749 | 861 | 710 | 680 | 894 | 327 | 419 |
| OM | 2491 | 2397 | 2367 | 2181 | 2181 | 1817 | 1897 | 1526 | 1049 | 1023 | 955 | 839 | 1286 | 668 | 540 |
| MH | 2661 | 2540 | 2514 | 2307 | 2307 | 1942 | 2058 | 1692 | 1206 | 1126 | 1094 | 950 | 1483 | 849 | 665 |

Table 14: Distance Matrix between the two NBA Conferences (inter-league).

For readability, the 30 × 30 distance matrix is broken into 15 × 15 matrices, providing both intra-league and inter-league distances. We remark that the Los Angeles Clippers and Los Angeles Lakers play their games in the same arena, which explains why their distance is zero. Each entry in Tables 12 through 14 is expressed in miles, unlike the NPB distance matrix which was expressed in kilometres.